\documentclass[final]{cvpr}
\pdfoutput=1
\usepackage{times}
\usepackage{epsfig}
\usepackage{graphicx}
\usepackage{amsmath}
\usepackage{amssymb}
\usepackage{comment}
\usepackage{multirow, multicol}
\usepackage[ruled,vlined]{algorithm2e}
\usepackage{comment}
\usepackage{subcaption}

\usepackage[pagebackref=true,breaklinks=true,colorlinks,bookmarks=false]{hyperref}

\def\cvprPaperID{10350} 
\def\confYear{CVPR 2021}

\begin{document}

\title{Lossy Event Compression based on Image-derived Quad Trees \\
and Poisson Disk Sampling}

\author{Srutarshi~Banerjee\quad Zihao W. Wang\quad Henry H. Chopp, \\
Oliver Cossairt\quad Aggelos K. Katsaggelos\\
Northwestern University\\
{\tt\small srutarshibanerjee2022@u.northwestern.edu}

}
\maketitle

\begin{abstract}
With several advantages over conventional RGB cameras, event cameras have provided new opportunities for
tackling visual tasks under challenging scenarios with fast motion, high dynamic range, and/or power constraint. Yet unlike image/video compression, the performance of event compression algorithm is far from satisfying and practical. The main challenge for compressing events is the unique event data form, i.e., a stream of asynchronously fired event tuples each encoding the 2D spatial location, timestamp, and polarity (denoting an increase or decrease in brightness). Since events only encode temporal variations, they lack spatial structure which is crucial for compression. To address this problem, we propose a novel event compression algorithm based on a quad tree (QT) segmentation map derived from the adjacent intensity images. The QT informs 2D spatial priority within the 3D space-time volume. In the event encoding step, events are first aggregated over time to form polarity-based event histograms. The histograms are then variably sampled via Poisson Disk Sampling prioritized by the QT based segmentation map. Next, differential encoding and run length encoding are employed for encoding the spatial and polarity information of the sampled events, respectively, followed by Huffman encoding to produce the final encoded events. Our Poisson Disk Sampling based Lossy Event Compression (PDS-LEC) algorithm performs rate-distortion based optimal allocation. On average, our algorithm achieves greater than 6$\times$  higher compression compared to the state of the art.
\end{abstract}

\section{Introduction}

Inspired by biological visual systems, event cameras are novel sensors designed to capture visual information with a data form drastically different from traditional images and videos \cite{dvs640,dvs128}. The event pixels do not directly output the intensity signals as traditional cameras do. Instead, each pixel compares the difference between the current log-intensity state and the previous state, and fires an event when the difference exceeds the firing positive or negative thresholds. This sensing mechanism provides several benefits. First, event pixels operate independently which enables very low latency ($\sim10\mu$s) and therefore high speed imaging. Second, event cameras have high dynamic range (HDR, $\sim120$dB) compared to regular frame-based cameras ($\sim60$dB). Third, the events reduce redundant captures of static signals. Last, event cameras consume lower power (10mW) than traditional cameras ($\sim1$W). As such, event cameras have brought new solutions to many classical as well as novel problems in computer vision and robotics, including high frame-rate video reconstruction
 \cite{ed-vfs,scheerlinck2020fast,shedligeri2019photorealistic}, with HDR \cite{eventHDR2019,rebecq2019high} and high resolution \cite{wang2020eventsr,choi2019learning,gef}, and 3D reconstruction of human motion \cite{xu2019eventcap} and scenes \cite{rebecq2018emvs,kim2016real}, as well as odometry \cite{censi2014low, vidal2018ultimate} and tracking \cite{zhu2017tracking, lagorce2014asynchronous}.

\begin{figure*}[h!]
\begin{center}
\includegraphics[ height=110pt]{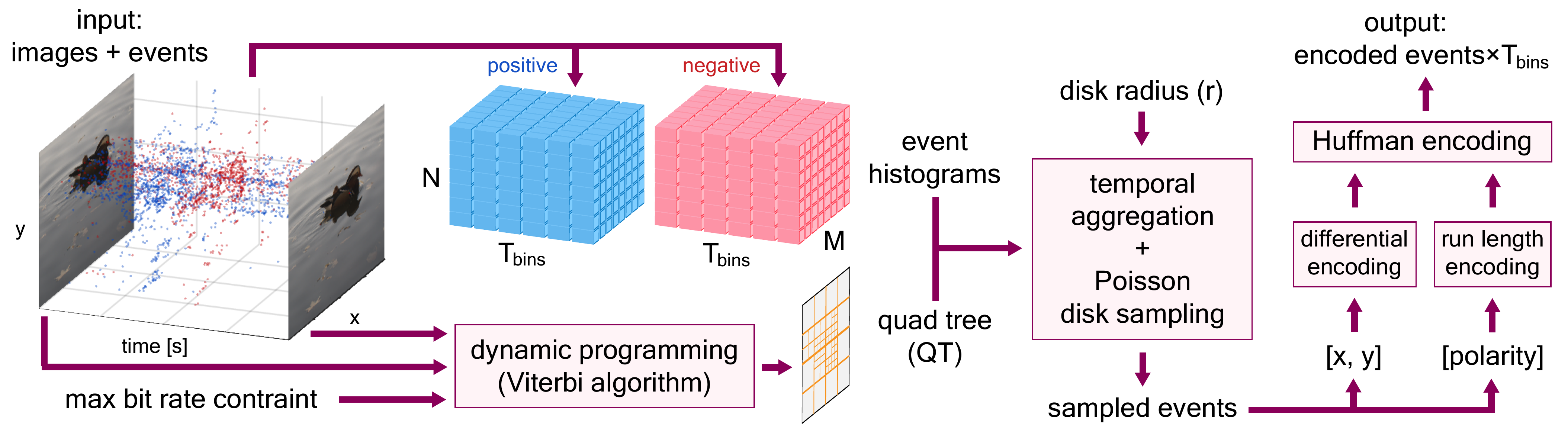}
\end{center}
   \caption{Overall Architecture of the PDS-LEC}
\label{fig: overall_architecture}
\end{figure*}

Currently, events are mainly communicated in the raw format using the Address Event Representation (AER) protocol \cite{AER_website}. The current AER protocol, AEDAT $4.0$, released in July 2019, \cite{AER_website} uses 96 bits representation for each event tuple (x, y, t, p) which are $(x, y)$ position, timestamp and polarity, while its earlier version, AEDAT $3.1$, uses 64 bit representation for each event. The timestamp uses the most bits due to its resolution with $64$ bits and $32$ bits for AEDAT $4.0$ and AEDAT $3.1$, respectively. Although AEDAT $4.0$ has incorporated lossless encoding standards such as LZ4, LZ4{\_}HIGH, ZSTD and ZSTD{\_}HIGH \cite{AER_website}, effective lossy event encoding has not been proposed in the literature or implemented in the event cameras. 

In traditional image/video compression standards, lossy compression is achieved by exploiting the spatial and temporal correlations. However, events are discrete points scattered in the space-time volume (see, for example Fig. \ref{fig: overall_architecture} left). Several prior works have approached event compression \cite{bi2018spike, dong2018spike, TALVEN}. TALVEN \cite{TALVEN} has aimed at aggregating events based on event timestamps. While this improves the compression ratio (CR), the benefits of high compression is only evident when aggregating events over a long time duration (e.g., $> 20 ms$), which in turn reduces the advantages of events. 

Our goal is to design an efficient event compression algorithm that exploits the spatio-temporal redundancy of events without sacrificing temporal information. Our approach considers the hybrid input of RGB images and events, which is a context largely existing in the literature \cite{zhu2018multivehicle, davis, gef, vidal2018ultimate}. We reason that the adjacent in time images have rich spatial features that can be leveraged to guide event compression. Therefore, we first compute a quad tree (QT) structure to serve as the priority map. Note that the QT can be generated from events as well. Next, the events are sampled to generate a blue noise distribution locally with the QT defining regions of importance, albeit in 2D space. The blue noise pattern of events is generated by Poisson Disk Sampling (PDS) \cite{bridson2007fast} which randomly samples events. The overall framework is shown in Fig. 1. In particular, this paper makes the following contributions: 

\emph{(1)} We propose a novel algorithm for lossy compression of neuromorphic events based on the QT segmentation map from adjacent images exploiting spatio-temporal redundancy.

\emph{(2)} We evaluate our approach qualitatively and quantitatively, on existing event datasets and compare our results with state of art and other techniques.

\section{Related Work}

Conventional image and video based coding techniques have evolved over the last couple of decades which led to a successful telecom revolution globally. Several standards came out as a result of the persistent effort, some of the latest video standards being HEVC, H.264, AVS2, VP9 \cite{HEVC_overview, H_264, AVS2_overview, VP9_overview}. With images and videos, the 2D and/or 3D spatio-temporal consistency of the data is exploited to have an efficient coding algorithm. However, events are asynchronous and non-continuous in space and time. Limited work \cite{dong2018spike, bi2018spike} has been done in predicting the distribution of events in space, time and polarity.

\subsection{DVS Coding Approaches}
The approaches for compressing DVS data can be classified into two categories: (a) specifically encoding events, (b) applying existing methods to events.

\emph{\textbf{A. DVS Specific Coding}}

\emph {(1) Spike Coding}: Lossless event compression was first proposed in \cite{bi2018spike}. The encoding was derived from a spike firing model exploiting spatial and temporal correlation. The events are projected into a sequence of three-dimensional macro-cubes and encoded using Address-Prior (AP) and Time-Prior (TP) modes. The AP mode is designed for scenarios where events are scattered all over the 2D space, while the TP mode is designed for cases where events occur locally. The residuals are encoded using arithmetic coding (CABAC). The authors extended this algorithm to include intercube prediction utilizing the temporal correlation among microcubes \cite{dong2018spike}, although the compression ratio did not improve significantly.  

\emph {(2) TALVEN}: The time aggregation strategy based lossless video encoding strategy was devised in \cite{TALVEN}. The events are time-aggregated into an event frame with a histogram count at each pixel location where each event frame is separated based on event polarities. The concatenated frames are coded using conventional lossless video encoding techniques including interframe coding, intraframe coding and entropy coding (CABAC) with the data packets incorporated into Network Abstraction Layer (NAL) units. The CR of this technique is significantly higher than the Spike Coding especially at higher time aggregation.   

\emph{\textbf{B. Existing compression methods for Events}}

The application of existing compression methods tailored to events can be broadly categorized into:

\emph{(1) Entropy coding} - 
Entropy coders like Huffman and arithmetic coding can potentially be applied to events \cite{event-survey} considering each event field $(x,y,t,p)$ as separate symbols. The gains of using just this technique are rather limited.

\emph{(2) Dictionary based compression} - A collection of long strings and shorter codewords is maintained in a dictionary data structure. Advanced dictionary coders Zstd \cite{Zstandard}, Zlib \cite{zlib}, LZ77 \cite{LZ77}, LZMA \cite{LZMA} and Brotli \cite{brotli} utilize multi-level encoding to improve compression ratios.

\emph{(3) IoT specific compression} - An Internet of Things (IoT)-specific strategy has been devised for resource constrained devices \cite{sprintz} without violating memory and latency constraints of the IoT devices. Sprintz exploits correlations among successive samples of a multivariate stream. The event stream can be directly converted to a multivariate time series data for compression. This strategy works better when it is required to have higher compression gains with reduced power consumption \cite{event-survey}.

\emph{(4) Fast integer compression} - In order to boost up the encoding and decoding speed (e.g. billions of arrays of integers for search engines), fast integer compression is often the choice. Simple8B \cite{sprintz}, Memcpy \cite{sprintz}, SIMD-BP128 \cite{simdbp128}, FastPFOR \cite{simdbp128} and SNAPPY \cite{snappy} are some of the fast integer compression algorithms. When DVS data needs a rapid data transfer, fast integer compression can be applied by transforming events into vectors.

A comparison of these approaches has been done in \cite{khan_comparison, TALVEN} resulting in low compression ratios between 1 and 4. The aforementioned approaches do not consider spatio-temporal encoding of events at the same time.

\subsection{Blue Noise Sampling}
Blue noise based sampling is widely applied in rendering, texturing, animation and related domains. Blue noise captures the local density of the image through its local point density. It has isotropic properties which lead to high-quality sampling of multidimensional signals without any aliasing \cite{2020blue}.  Blue noise distribution has been generated over the years by several techniques--error diffusion \cite{digital_half, floyd1976}, dart throwing algorithm \cite{cook1986}--to create Poisson disk distributions and other techniques. Poisson disk algorithms that generate blue noise samples \cite{BlueNoise_survey, BlueNoise1, BlueNoise2, BlueNoise3, BlueNoise4, BlueNoise5} by reducing the computational complexity and improving the blue noise characteristics are widely studied and applied in the field of computer graphics. 

We leverage a fast Poisson Disk Sampling (PDS) technique as a method to sample events in 2D space. This generates a blue noise pattern preserving the high frequency spectral components of the events. We compare PDS with random sampling in Section $6$ and show that using PDS results in overall higher compression ratio and better event compression metrics.

\section{Spatio-Temporal Event Compression}
Mathematical analysis in \cite{bi2018spike, dong2018spike} points out the correlation of DVS spikes in both space and time. The adjacent pixels receive the same illumination change (increase or decrease) over time. This leads to spatial as well as temporal correlation in local regions. We exploit this fundamental nature of the DVS spikes for encoding events. First, we generate a quad tree (QT) structure from the adjacent intensity frame which indicates spatial priority regions. Depending on the block sizes, we develop a strategy based on Poisson disk sampling for removing events based on spatio-temporal proximity to each other. The events are encoded based on block sizes using entropy encoding strategies. In the following subsections, the framework is described in more details.

\subsection{Generation of Quadtree (QT) structure}
The events are generated due to scene complexity (edges/textures of the objects in scene) and (or) relative motion of the camera with respect to the scene. Khan \etal \cite{khan_BW_2019} showed that the event rate depends exponentially on the scene complexity metric and linearly on the sensor speed metric. However, the scene complexity provides us with the important events pertaining to objects.

We rely on the intensity frames to generate the priority regions for the events. Although priority can be generated in different ways, in this work we use the QT data structure in order to divide the frame into blocks. The QT can subdivide the frames into different levels resulting in heterogeneous blocks of sizes determined by the leaf level of the QT. Larger block sizes indicate lower priority regions while smaller blocks indicate higher priority. A time aggregated event stream between two successive intensity frames can be approximated by a difference of these intensity frames. Hence, the priority regions for events is proportional to the difference between these intensity frames. We generate the QT based on the system described in \cite{Banerjee_EUSIPCO_2019}. For an event volume $E_{t-1}$, between successive intensity frames $\hat{I}_{t-1}$ and $I_t$ (both as defined in \cite{Banerjee_EUSIPCO_2019}), we leverage on these intensity frames to derive a relevant QT structure. The QT is generated using Dynamic Programming (Viterbi optimization \cite{bezier_sch}) which provides a trade-off between the frame distortion ($D$) and frame bit rate ($R$). This is done by minimizing $D$ over the leaves of the QT, denoted by $\textbf{x}$, subject to a given maximum bit rate $R_{max}$. In other words, we are solving for the rate-distortion tradeoff for the intensity frames. The constrained discrete optimization is solved using Lagrangian relaxation, leading to solutions in the convex hull of the rate-distortion curve. The Lagrangian cost function is of the form
\begin{align}
\label{eqn:eqlabel1}
    J_{\lambda} (\textbf{x}) = D(\textbf{x})+ \lambda R(\textbf{x}),
\end{align}
where $\lambda \geq 0$ is a Lagrangian multiplier. 
The intensity bit rate is set to be generated for a fixed rate (within a tolerance) constraint. The QT is generated such that the intensity bit rate satisfies this constraint. $\lambda$ is adjusted at each frame for achieving the desired bit rate. The optimal $\lambda^{*}$ is computed by a convex search in the Bezier curve \cite{bezier_sch} over the rate-distortion space which results in convergence in fewer iterations. The optimal QT segmentation map corresponds to $\lambda^{*}$ and at a set bit rate, satisfies equation  (\ref{eqn:eqlabel1}), derived for intensity frame $I_t$ based on the distorted frame $\hat{I}_{t-1}$. Clearly, at higher bit rates the QT can go to deeper levels (and hence smaller blocks) while for smaller bit rates, the QT branches out to shallow levels. This 2D segmentation map is applied to the event volume $E_{t-1}$ providing regions of priority for event compression, with block size inversely proportional to the priority in the block.

\subsection{Poisson Disk Sampling}
The QT structure provides a segmentation map. Since the events are correlated locally in space and time, there exists local redundancy not only in the temporal dimension but also in the spatial dimensions. The local region in space is quite abstract as it may vary from scene to scene and also within a scene. In this work, the QT block sizes indicate the importance of the regions, which is also an indication of the dimensions of local regions in space-time. There could be local features in bigger blocks, but these are not of high priority based on the set bit rate on intensity frames.

We perform lossy compression on events not only by sampling the events from the QT blocks, but also temporally aggregating the events. Sampling of events inside the QT blocks is challenging due to the presence of both local structure and noise. There may be aggregation of events at object edges, which may be increased due to presence of noise in the sensor. The events are sampled as a blue noise pattern. The blue noise pattern picks up the local event density without adding any inherent artificial structures. This leads to high-quality sampling of multi-dimensional signals while preventing aliasing. In this work, PDS is used to generate blue noise.

PDS is applied on the events inside the QT blocks. PDS sample events, with all events at least $r$ distance apart from each other. Bridson \cite{bridson2007fast} proposed a fast PDS strategy in multidimensional space. However, the algorithm would generate points in space based on the PDS. In our work, we sample events based on this PDS technique. The resulting sampled events are generated with at least $r$ distance apart from each other. Given $M$ original events, we sample $R_e$ events, where $R_e <= M$. PDS can be applied on the whole event volume or on regions of the events. Since the QT already provides us with a priority map, PDS is done with different $r$ on blocks of different sizes in the QT.  

PDS needs a reference position to start sampling the events. In a QT block, it is difficult to identify a reference starting point for PDS. The reference starting point can be conveniently in any of the corner locations in the QT block (or any other location in the QT block). However, this might lead to a reference point which may not have an event. In order to eliminate any such issues, we consider the reference point as the location of the event at or nearest to the centroid (geometric median) with respect to the event locations in a QT block. Thus, in a neighborhood of $N \times N$ pixels, we find the geometric median  $x_m$, of the $M$ events, as shown in Eqn. (2), where each $x_i \in \mathbb{R}^n$ are the event locations in space-time.
\begin{align}
\label{eqn:eqlabel2}
   x_{m} = \underset {x_{m}} {\text{arg
min}}\sum_{i=1}^{M} \| x_{i} - x_{m}\|_2,
\end{align}
Except for the chosen point, we do not sample the events lying within a disk of radius $r$, with respect to this reference point. Next, we shift our reference point to the nearest event location outside the radius $r$ and do not sample the events lying within radius $r$ (termed as Poisson Disk Radius (PDR)) with respect to this new reference point. This step is repeated until all the events are at least or greater than a spatial distance $r$ from each other. This sampling strategy has the effect of sampling the event density locally in the region of $N \times N$ pixels, along with introducing a fixed spatial sampling pattern in the event volume. The sampling approach not only reduces the density of the aggregated events near the edges but can also remove the uncorrelated noisy events from local regions in space.

\subsection{Lossy Event Compression}
The QT provides the priority map and PDS provides an efficient strategy to reduce the density of events and noise from the events. However, the QT has different sized blocks which indicate different priorities. We want to sample events based on priority as this will retain local features in high priority regions and remove local features from lower priority regions. This is achieved by using different Poisson disk radius $r$ values for different QT block sizes: higher and smaller $r$ values for bigger and smaller QT blocks, respectively. In our algorithm, we specifically apply PDS to blocks larger than $2 \times 2$. For QT block sizes below $4 \times 4$, we do not sample events as we consider those blocks to be regions of very high priority and hence all events in those regions are critical. As mentioned in subsection $3.1$, we obtain the QT for event volume $E_{t-1}$ from reconstructed intensity frame $\hat I_{t-1}$ and actual frame $I_{t}$.

\setlength{\textfloatsep}{7pt}
\begin{algorithm}[ht]
\KwIn{Adjacent Intensity frames, Event Volume}
\SetAlgoLined
 set $\lambda_{min}$, $\lambda_{max}$, $R_{max}$\;
  \While{intensity bit rate $R > R_{max}$}{
  adjust $\lambda$ to obtain desired $R$ as in Sec. $3.1$
  }
  \KwResult{Optimal QT}
 set $T_{bin}$ and $r_4$\;
 Aggregate events into event histogram frames (positive and negative) based on $T_{bin}$;
 
 \While{ QT blocks $\in$ \{ $4\times4$, $8\times8$ and higher \} }{
  compute the centroid $C$ of events in $(x,y)$\;
  Shift $C$ to nearest event location $P$ (if $C \neq P$) \;
  
  \If{events left to be visited}{
  Remove events within radius $r_i$ from $P$\;
  set $P$ as nearest event outside $r_i$\;
  }
  Encode $(x,y)$ differentially followed by Huffman coding for each block\;
  Encode event counts as Run Length Encoding each block followed by Huffman coding\;
 }
 
\While{ QT blocks $\in$ \{ $2\times2$ \} }{
 Encode $(x,y)$ differentially each block\;
 Encode event counts as Run Length Encoding each block followed by Huffman coding\;
}

\While{ QT blocks $\in$ \{ $1$ pixel \} }{
 Consider all pixels in the frame\;
 Encode event counts as Run Length Encoding followed by Huffman coding\;
}
 \KwOut{Compressed Event Volume}
 \caption{Event Compression Pseudo Code}
\end{algorithm} 

The lossy event compression is performed in steps depending on the bandwidth available for communication between network-connected IoT devices. The lossy event compression framework has a lossy part as well as a lossless part. The lossy portion of event encoding consists of 2 parts: (a) quantizing (aggregating) event timestamps as $T_{bin}$ and (b) PDS of the events with spatial overlap of event $(x,y)$ locations within the QT blocks. In step (a), we compute the histogram for the positive and negative events separately for each quantized timestamp. The lossless portion of the event encoding involves differentially encoding the $(x,y)$ location of events followed by Huffman coding. The polarity is encoded by Run Length Encoding (RLE) followed by Huffman encoding (HE). The quantized (aggregated) timestamps for the events form the basis of aggregating events as subframes corresponding to that timestamp. For each timestamp, there are 2 subframes - one for positive and one for negative events. These events are grouped together as a frame before transmission/storage. The algorithm is mentioned in detail in Algorithm 1 with overall architecture shown in Fig. $1$. It must be clearly stated that we are developing a heuristics based rule for compressing the events. This algorithm offers flexibility in terms of event compression. Depending on the desired bitrate, the event compression may be set by the user, by varying $r$ and $T_{bin}$. 
\section{Experiments}
\subsection{Dataset}
In order to show the benefits of the proposed lossy compression strategy, we use datasets available in the literature which have both intensity frames and events. Sequences from RGB-DAVIS dataset \cite{gef,gef-dataset} are used to show the compression performance in different settings for image reconstruction experiments. The Dynamic and Active-pixel Vision Sensor (DAVIS) dataset \cite{davis} is used to show a comparison of PDS-LEC with other state-of-the-art event compression algorithms. Shapes-6dof, Dynamic-6dof, Slider-depth, Outdoors Running sequence has been used in DAVIS dataset. Running1 and Running2 sequences has been derived from Outdoors Running as done in \cite{TALVEN}.

\subsection{Event Compression Metric}
\emph{\textbf{(a) Event Distortion Metric:}} 
The compressed events essentially represent the original 3D spatio-temporal event volume in a quantized manner. However, the fidelity of the compressed events with respect to undistorted event volume is very important to understand the level of distortion. In this direction, to the best of our knowledge, there is no metric for computing event distortion and we are making a first attempt in quantizing the distortion. The event spatial location $(x, y)$ is not only important, but the timestamp is also a vital parameter for various event based processing algorithms. The distortion of the events in both the spatial and temporal aspects must be computed.   

We separate the spatial and temporal fidelity of the distorted event volume with respect to the undistorted events in order to have a complete understanding of the role of different parameters in encoding events. For computing the spatial distortion, we aggregate the events over the time bins into an $(x,y)$ event image. The PSNR and SSIM metrics computed on the aggregated compressed and uncompressed event images are used for computing the spatial distortion. The temporal distortion is rather hard to quantify. We define a temporal error metric to quantify the quantization error in time as described in Eqn. $3$:
\begin{align}
\label{eqn:eqlabel7}
    T_{error}=\frac{1}{N_{fr}}{\sum_{i=1}^{N_{fr}}\sqrt{\sum_{j}{\left( T_{j,org} - T_{j,quant} \right)^2}}},
\end{align}
where $N_{fr}$ is the number of event volumes in a sequence, $T_{j,org}$ is the timestamp of $j^{\text{th}}$ event in $i^{\text{th}}$ frame, and $T_{j,quant}$ is the quantized timestamp of $j^{\text{th}}$ event in a compressed $i^{\text{th}}$ frame. Based on the end application of events, a weighted distortion metric may be computed with weights $w_{s}$ and $w_{t}$ for spatial distortion $D_{s}$ and temporal distortion $D_{t}$, respectively, thereby prioritizing spatial and / or temporal distortions differently.

\emph{\textbf{(b) End-to-End Event Compression Metric:}} We also compute the end-to-end compression ratio for the events \cite{TALVEN}, considering 64 bits for each event. The compression ratio is given by the ratio of uncompressed events and the encoded event bit representation.
\begin{align}
\label{eqn:eqlabel9}
    Ev_{comp} = \frac{{Bits/event} \times N_{events}}{\gamma} ,
\end{align}
where $\gamma$ is the compressed event bits.

\section{Performance of proposed framework}
Here we show a performance study of the proposed lossy event compression framework with respect to different compression parameters.

\subsection{Performance with varying $T_{bin}$ only}
In one of the experiments, the QT segmentation is not used. The events are not sampled based on Poisson disk sampling. However, the events are temporally aggregated into $N$ bins, with $N \in \{8,16,24\}$. To quantify the benefits of the temporal compression, we compute the compression ratio of the original uncompressed events as well as the compressed events. Table \ref{tab:temporal binning only} shows the compression ratio and the average $T_{error}$ over the sequence.
\begin{table}[htbp]
    \centering
    \resizebox{\linewidth}{!}{\begin{tabular}{|c| c c c| c c c|}
        \hline
        ~ & \multicolumn{3}{c |}{$T_{error}$} & \multicolumn{3}{c |}{CR (bit rate ratio)} \\
        \hline 
        Sequence & 8 & 16 & 24 & 8 & 16 & 24 \\
        \hline
        Indoor3 & 0.269 & 0.134 & 0.089 & 1029.63 & 944.44 & 890.41\\
        Indoor4 & 0.139 & 0.069 & 0.046 & 279.34 & 263.13 & 251.18 \\
        Indoor6 & 0.278 & 0.139 & 0.093 & 1643.04 & 1500.79 & 1371.94 \\
        Indoor9 & 0.341 & 0.166 & 0.113 & 3331.48 & 2643.58 & 2404.56 \\
        \hline
        Outdoor5 & 0.149 & 0.076 & 0.051 & 41.84 & 40.68 & 39.65 \\
        Outdoor6 & 0.256 & 0.129 & 0.086 & 514.35 & 490.17 & 466.52 \\
        Outdoor9 & 0.183 & 0.089 & 0.062 & 45.03 & 44.23 & 43.46 \\
        \hline
    \end{tabular}}
    \caption{Temporal binning only. CR: compression ratio}
    \label{tab:temporal binning only}
\end{table}
Clearly, it can be seen that with the increase in the number of $T_{bin}$, the compression ratio reduces along with the reduction of $T_{error}$. Fig. $2 (a)$ shows the original events while  Figs. $ 2 (b), (c)$ and $(d)$ show the variation of events in temporal space with 8, 16 and 24 quantized timestamps respectively. The SSIM in this case is 1, since all events are sampled. 

\subsection{Performance with varying Intensity Bit Rate}
The QT is optimized for particular operational bit rate for intensities only. In these experiments, $r_4 = 1$, $r_8 = 2r_4$, $r_{16} = 3r_4$ and $r_{32} = 4r_4$ with $T_{bin} = 16$ is used. The performance table is shown in Table \ref{tab:varying bit rate}. As bit rate reduces, the PSNR and SSIM reduces, while $T_{error}$ and CR increases. This indicates bigger blocks in the QT for lower bit rates with considerably higher distortion. Sample event frames at varying intensity bit rate is shown in Fig. $2$ (row $2$).
\begin{table}[htbp]
    \centering
    \resizebox{\linewidth}{!}{\begin{tabular}{|c|c|c|c|c|c|}
        \hline
        Sequence & Bit Rate (Mbps) & PSNR (dB) & SSIM & $T_{error}$ & CR (bit rate)\\
        \hline
        \multirow{3}{*}{Indoor3} & 0.5 & 43.07 & 0.9999 & 0.177  & 994.41 \\
         & 0.3 & 41.71 & 0.9999 & 0.197 & 1048.16 \\
         & 0.1 & 38.97 & 0.9999 & 0.200 & 1447.97 \\
        \hline
        \multirow{4}{*}{Indoor4} & 0.5 & 41.38 & 0.9983 & 0.233  & 271.68 \\
         & 0.3 & 40.33 & 0.9977 & 0.260 & 279.67 \\
         & 0.1 & 25.60 & 0.9999 & 0.515 & 1797.60 \\
        \hline
        \multirow{3}{*}{Indoor6} & 0.5 & 39.32 & 0.9998 & 0.205  & 1639.61 \\
         & 0.3 & 39.00 & 0.9998 & 0.220 & 1655.10 \\
         & 0.1 & 36.82 & 0.9996 & 0.267 & 1805.60 \\
        \hline
        \multirow{3}{*}{Indoor9} & 0.5 & 31.85 & 0.9970 & 0.336 & 2053.90 \\
         & 0.3 & 31.42 & 0.9961 & 0.356 & 2052.87 \\
         & 0.1 & 30.43 & 0.9936 & 0.400 & 2388.33 \\
        \hline
        \multirow{3}{*}{Outdoor5} & 0.5 & 33.90 & 0.9895 & 0.510  & 47.63 \\
         & 0.3 & 32.70 & 0.9868 & 0.566 & 50.30 \\
         & 0.1 & 31.37 & 0.9778 & 0.684 & 59.97 \\
        \hline
        \multirow{3}{*}{Outdoor6} & 0.5 & 29.86 & 0.9999 &   0.332 & 562.84 \\
         & 0.3 & 27.69 & 0.9998 & 0.388 & 603.45 \\
         & 0.1 & 25.16 & 0.9997 & 0.456 & 713.15 \\
        \hline
        \multirow{3}{*}{Outdoor9} & 0.5 & 30.99 & 0.9935 &  0.5580 & 52.90 \\
         & 0.3 & 30.08 & 0.9919 & 0.6093 & 56.49 \\
         & 0.1 & 28.75 & 0.9879 & 0.7055 & 65.89 \\
        \hline
    \end{tabular}}
    \caption{Performance with varying Bit Rate ($r_4 = 1$)}
    \label{tab:varying bit rate}
\end{table}
\begin{table}[htbp]
    \centering
    \resizebox{\linewidth}{!}{\begin{tabular}{|c|c|c|c|c|c|}
        \hline
        Sequence & PDR ($r_4$) & PSNR (dB) & SSIM & $T_{error}$ & CR (bit rate)\\
        \hline
        \multirow{3}{*}{Indoor3} & 1 & 41.71 & 0.9999 & 0.197 & 1048.16\\
         & 2 & 39.52 & 0.9999 & 0.230 & 1117.74\\
         & 3 & 37.91 & 0.9999 & 0.257 & 1158.69\\
        \hline
        \multirow{4}{*}{Indoor4} & 1 & 40.33 & 0.9977 & 0.260 & 279.67\\
         & 2 & 37.16 & 0.9954 & 0.348 & 290.70\\
         & 3 & 35.22 & 0.9939 & 0.398 & 298.73\\
        \hline
        \multirow{3}{*}{Indoor6} & 1 & 39.00 & 0.9998 & 0.220 & 1655.10\\
         & 2 & 36.70 & 0.9996 & 0.276 & 1742.88\\
         & 3 & 35.45 & 0.9995 & 0.307 & 1807.72\\
        \hline
        \multirow{3}{*}{Indoor9} & 1 & 31.42 & 0.9961 & 0.356 & 2052.87\\
         & 2 & 29.49 & 0.9922 & 0.424 & 2117.61\\
         & 3 & 28.38 & 0.9899 & 0.461 & 2159.61\\
        \hline
        \multirow{3}{*}{Outdoor5} & 1 & 32.70 & 0.9868 & 0.566 & 50.30\\
         & 2 & 30.82 & 0.9741 & 0.712 & 60.40\\
         & 3 & 30.02 & 0.9661 & 0.783 & 69.21\\
        \hline
        \multirow{3}{*}{Outdoor6} & 1 & 27.69 & 0.9998 & 0.388 & 603.45\\
         & 2 & 23.77 & 0.9997 & 0.488 & 712.75\\
         & 3 & 20.96 & 0.9996 & 0.549 & 794.22\\
        \hline
        \multirow{3}{*}{Outdoor9} & 1 & 30.08 & 0.9919 & 0.609 & 56.49\\
         & 2 & 27.32 & 0.9819 & 0.801 & 72.00\\
         & 3 & 25.93 & 0.9742 & 0.898 & 86.39\\
        \hline
    \end{tabular}}
    \caption{Performance with varying $r_4$ at 0.3 Mbps}
    \label{tab:varying r4}
\end{table}

\begin{figure*}
\begin{center}
\includegraphics[width=.85\linewidth]{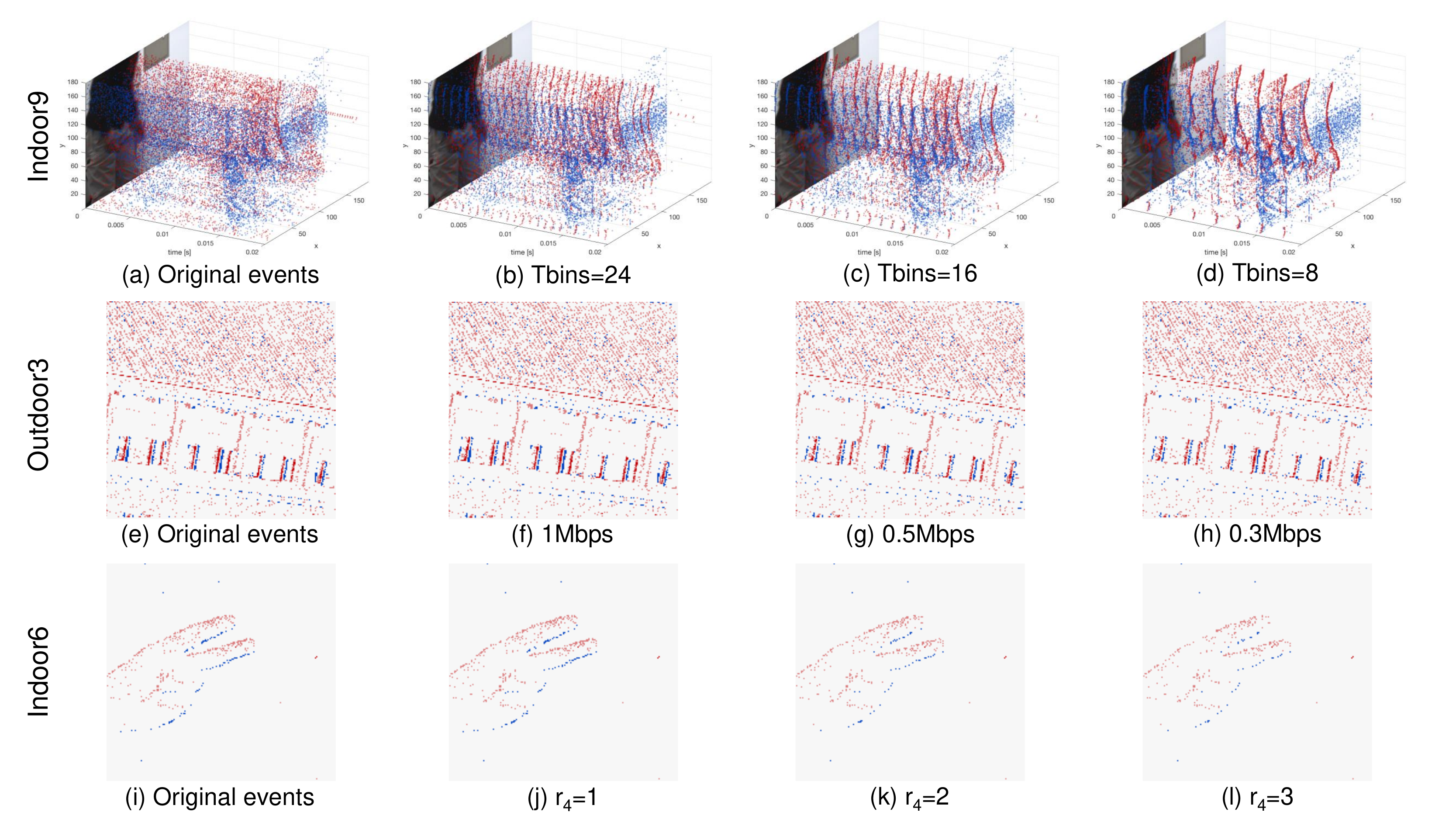}
\end{center}
   \caption{Row 1: temporal binning comparison for Indoor9 (human walking); Row 2: bit rate comparison for Outdoor3 (building) with $r_{4}=1$; Row 3: $r_4$ comparison for Indoor6 (hand gesture).}
\label{fig:evaluation}
\end{figure*}
\begin{figure*}
\begin{center}
\begin{subfigure}{0.32\linewidth}
\includegraphics[height=110pt]{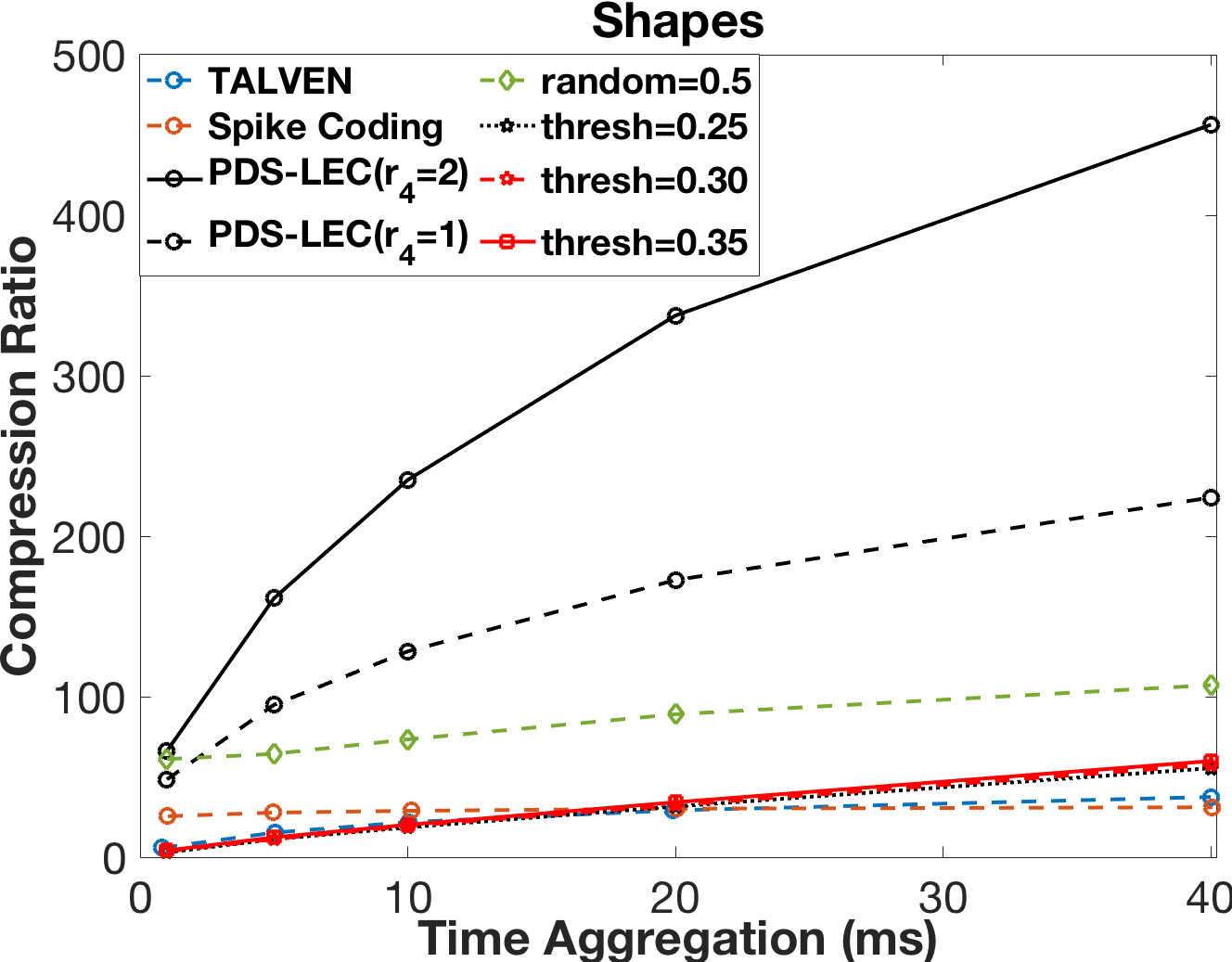}
\caption{Shapes sequence}
\end{subfigure}
\begin{subfigure}{0.32\linewidth}
\includegraphics[height=110pt]{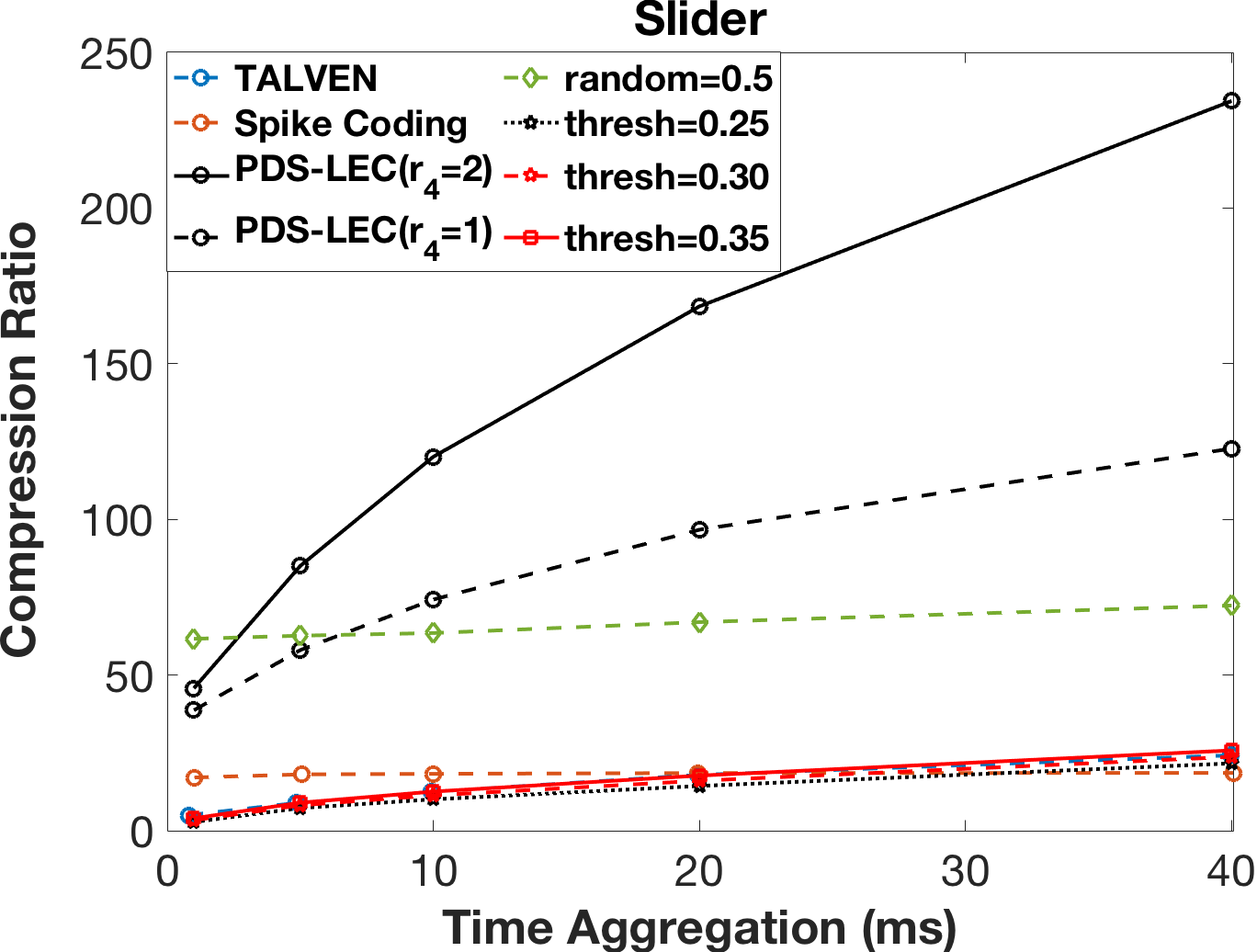}
\caption{Slider sequence}
\end{subfigure}
\begin{subfigure}{0.32\linewidth}
\includegraphics[height=110pt]{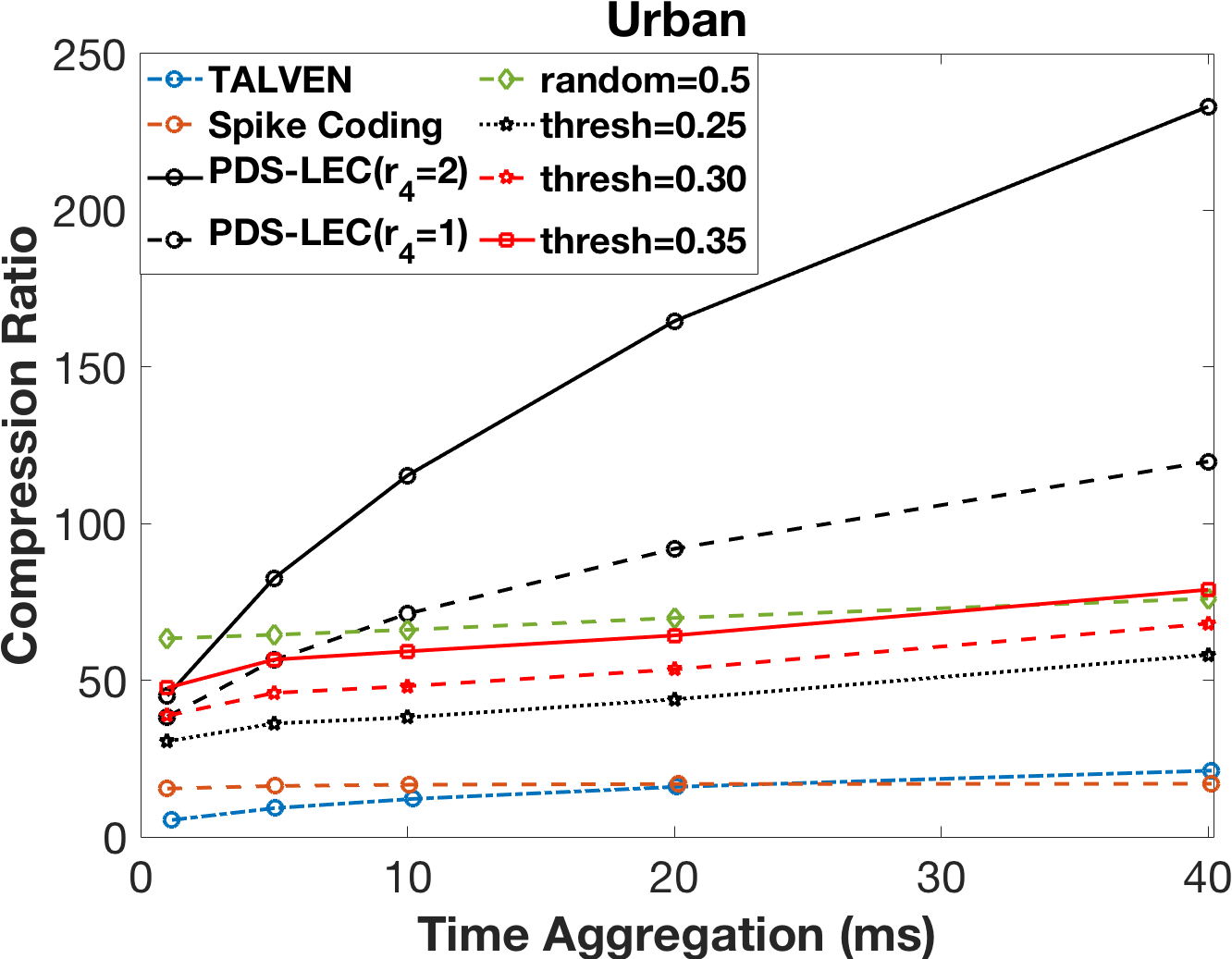}
\caption{Urban sequence}
\end{subfigure}
\caption{Compression Ratio of PDS-LEC (our work) at $100$ kbps, $r_4 = 1, 2$, with TALVEN \cite{TALVEN}, Spike Coding \cite{dong2018spike}, random (random $= 0.5$) and CT (thresh $= 0.25, 0.30, 0.35$) techniques}
\label{fig:comparison_100kbps}
\end{center}
\end{figure*}
\begin{figure*}
\begin{center}
\begin{subfigure}{0.32\linewidth}
\includegraphics[height=110pt]{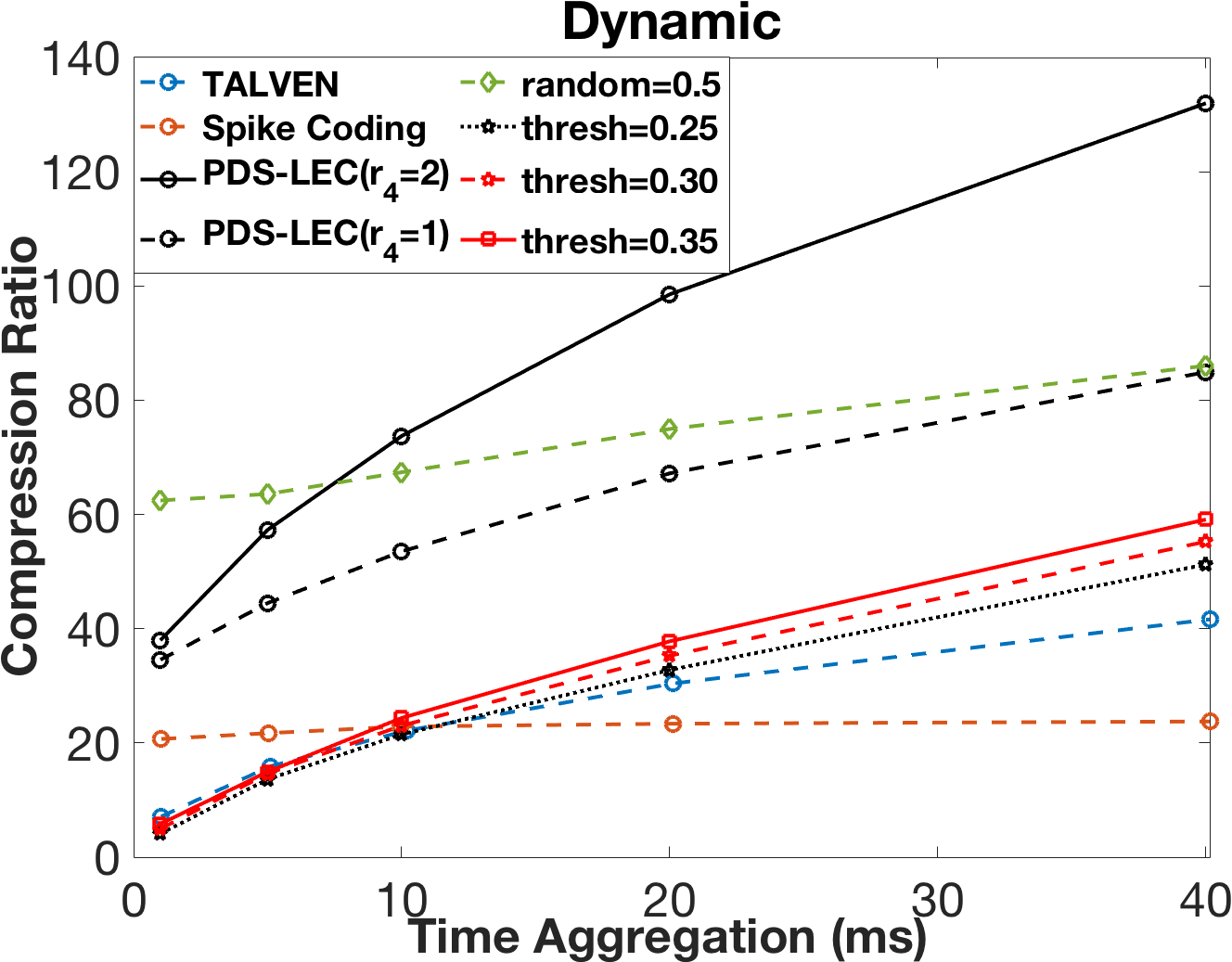}
\caption{Dynamic sequence}
\end{subfigure}
\begin{subfigure}{0.32\linewidth}
\includegraphics[height=110pt]{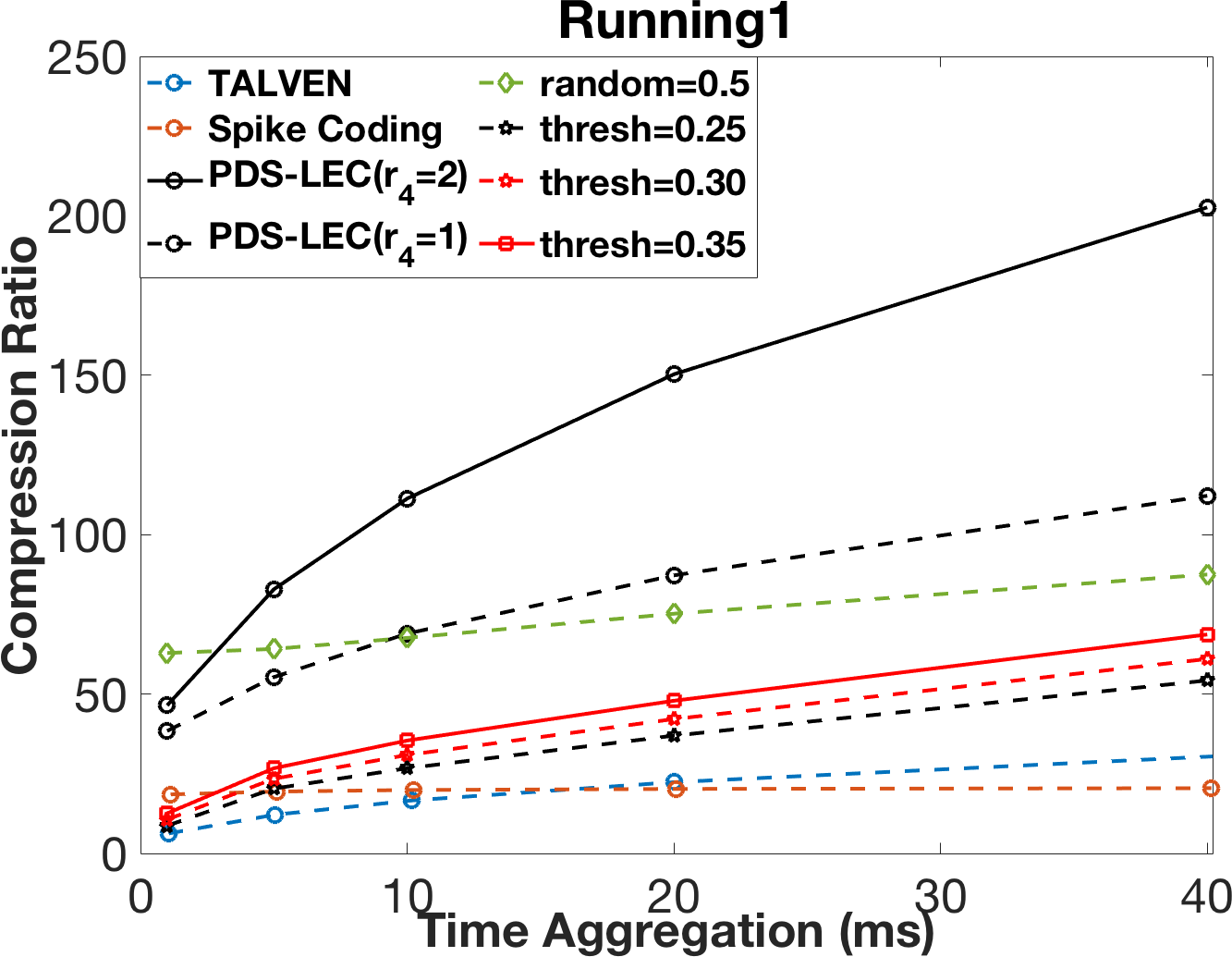}
\caption{Running1 sequence}
\end{subfigure}
\begin{subfigure}{0.32\linewidth}
\includegraphics[height=110pt]{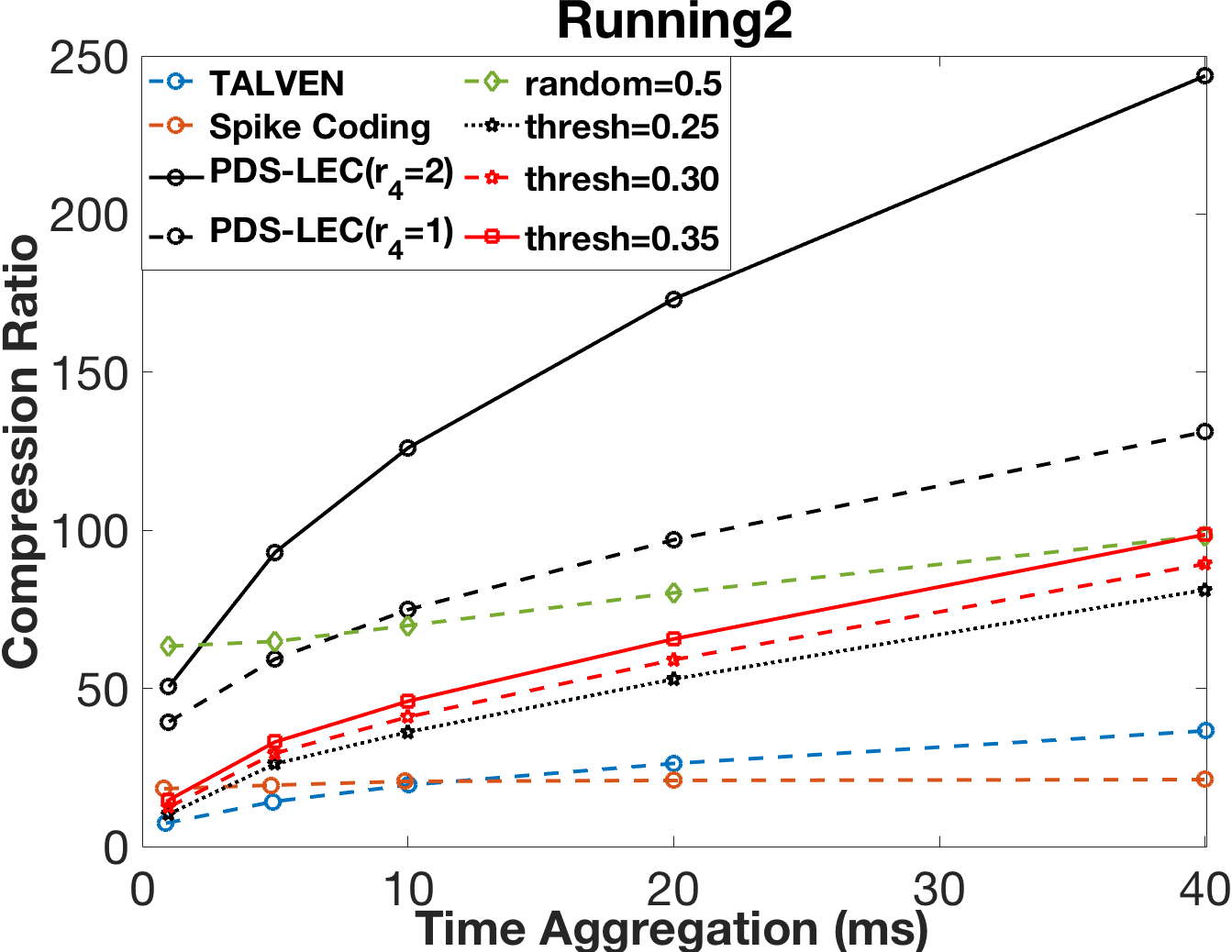}
\caption{Running2 sequence}
\end{subfigure}
\end{center}
\caption{Compression Ratio of PDS-LEC (our work) at $300$ kbps, $r_4 = 1, 2$, with TALVEN \cite{TALVEN}, Spike Coding \cite{dong2018spike}, random (random $= 0.5$) and CT (thresh $= 0.25, 0.30, 0.35$) techniques}
\label{fig:comparison_300kbps}
\end{figure*}
\begin{figure}
\begin{center}
\begin{subfigure}{0.49\linewidth}
\includegraphics[width=\textwidth]{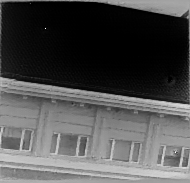}
\caption{Original \\reconstruction}
\end{subfigure}
\begin{subfigure}{0.49\linewidth}
\includegraphics[width=\textwidth]{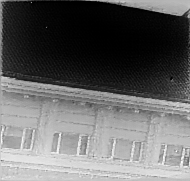}
\caption{Tbins = 24 (PSNR = 18.46, SSIM = 0.7449)}
\end{subfigure}
\begin{subfigure}{0.49\linewidth}
\includegraphics[width=\textwidth]{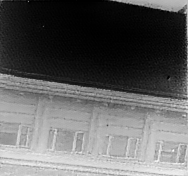}
\caption{Tbins = 16 (PSNR = 16.97, SSIM = 0.3354)}
\end{subfigure}
\begin{subfigure}{0.49\linewidth}
\includegraphics[width=\textwidth]{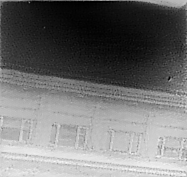}
\caption{Tbins = 8 (PSNR = 13.63, SSIM = 0.2337)}
\end{subfigure}
\end{center}
   \caption{Image reconstruction with encoded events with time duration $33$ ms, (Outdoor 3 sequence, Frame 6)}
\label{fig:application}
\end{figure}

\subsection{Performance with varying Poisson Disk Radius and Temporal Binning}
In this experiment, the performance of the event compression system is evaluated at a particular bit rate, but the Poisson disk radius $r_4$ is varied. As in section $5.2$, $r_8 = 2r_4$, $r_{16} = 3r_4$ and $r_{32} = 4r_4$. The temporal binning of $T_{bin} = 16$ is used in this experiment. The PSNR and SSIM reduces as $r_4$ increases as shown in Table \ref{tab:varying r4}. The $T_{error}$ increases with an increase of $r_4$, indicating the increase of temporal distortion. The compression ratio also increases with an increase of $r_4$. Sample event images for the original and compressed events for the Indoor6 sequence (frame 150) is shown in Fig. $2$ (row 3)  at 100 kbps. With the increase in $r_4$, the event image becomes less dense. It is noted that by setting different values of $r_4$, we can obtain a target bit rate for the events at a cost of a particular distortion.

\section{Comparative Performance measure with other benchmark strategies}
The proposed PDS-LEC algorithm is compared with other techniques such as TALVEN \cite{TALVEN} and Spike Coding \cite{dong2018spike} using sequences from DAVIS dataset \cite{davis}. In literature, they are the best performing event compression algorithms. In TALVEN, the events are aggregated into positive and negative event frames with histogram count at each pixel location, which are then coded (refer to subsection $2.1$). The compression ratio of other algorithms \cite{brotli, LZMA} in literature are even lower than TALVEN and Spike Coding \cite{TALVEN} and hence not included in the comparative study. Figs. $3$ and $4$ show the comparative results with different time aggregations. We also compare our method by replacing the QT with $16\times16$ uniform blocks and PDS by random sampling of $50\%$ of the events followed by temporal aggregation. The rest of the coding strategy in PDS-LEC is followed. Additional comparison is performed by generating fewer events for these sequences by using contrast thresholds (CT) of $0.25$, $0.30$ and $0.35$ in ESIM \cite{rebecq2018esim} and apply PDS-LEC by replacing QT with uniform $16\times16$ blocks and removing PDS step. PDS-LEC outperforms current existing algorithms for different bit rates and PDRs in terms of CR. Even for fewer events generated with higher CT (using our compression scheme), PDS-LEC has better CR. However, for 50\% random events (using our compression scheme), at lower temporal aggregation, in a few experiments, CR is higher than PDS-LEC, especially at QT derived at higher bit rates and/or lower PDR since less events are removed. However, SSIM and T$_{error}$ metrics as shown in Table \ref{tab:lossy_technique_comparison} indicate performance comparable or worse than PDS-LEC, due to random removal of events resulting in poor SSIM and $T_{error}$ compared to PDS-LEC  (please see additional details in supplementary material). Overall improvement in performance with other state-of-art and other alternative lossy techniques is due to temporal aggregation and priority based sampling of events. QT acts as a priority map for the events. The aggregated timestamps are converted into time frames which result is huge compression. Additionally, differential encoding, HE and RLE further increases CR. 
\begin{table}[htbp]
    \centering
    \resizebox{\linewidth}{!}{\begin{tabular}{|c|c|c|c|c|c|c|}
        \hline
        & & \multicolumn{5}{|c|}{SSIM, $T_{error}$}\\
        \hline
        Sequence & method & 1 ms & 5 ms & 10 ms & 20 ms & 40 ms\\
        \hline
        \multirow{6}{*}{Dynamic} & Rn & 0.850, 2.211 & 0.848, 2.222 & 0.843, 2.248 & 0.837, 2.406 & 0.831, 3.128\\
        & PL1 & 0.986, 0.956 & 0.937, 1.698 & 0.905, 1.978 & 0.871, 2.335 & 0.845, 3.128\\
        & PL2 & 0.969, 1.299 & 0.884, 2.099 & 0.840, 2.345 & 0.801, 2.609 & 0.772, 3.130\\
        & TH25 & 0.545, 2.409 & 0.545, 2.183 & 0.545, 2.070 & 0.545, 2.228 & 0.545, 3.127\\
        & TH30 & 0.558, 2.449 & 0.558, 2.229 & 0.558, 2.122 & 0.558, 2.257 & 0.558, 3.127\\
        & TH35 & 0.567, 2.551 & 0.567, 2.266 & 0.567, 2.165 & 0.567, 2.285 & 0.567, 3.128\\
        \hline
        \multirow{6}{*}{Running1} & Rn & 0.836, 2.467 & 0.834,  2.461 & 0.831, 2.467 & 0.825, 2.648 & 0.818, 3.491\\
         & PL1 & 0.968, 1.444 & 0.894, 2.234 & 0.852, 2.499 & 0.816, 2.802 & 0.790, 3.491\\
         & PL2 & 0.933, 1.930 & 0.819, 2.701 & 0.772, 2.911 & 0.737, 3.108 & 0.712, 3.490\\
        & TH25 & 0.554, 3.213 & 0.554, 2.841 & 0.554, 2.593 & 0.554, 2.619 & 0.554, 3.491\\
        & TH30 & 0.574, 3.235 & 0.574, 2.926 & 0.574, 2.684 & 0.574, 2.684 & 0.574, 3.491\\
        & TH35 & 0.588, 3.259 & 0.588, 2.990 & 0.588, 2.764 & 0.588, 2.745 & 0.588, 3.491\\
        \hline
    \end{tabular}}
    \caption{Comparison of lossy compression techniques. Rn: Random sampling of 50 $\%$ events. PL1, PL2: PDS-LEC with $r_{4}=1, 2$ at 0.3 Mbps. TH25, TH30, TH35: Fewer events with CT $=0.25, 0.30, 0.35$ respectively.}
    \label{tab:lossy_technique_comparison}
\end{table}

\section{Applications to Image Reconstruction}
\label{sec:appl}
We used E2VID \cite{rebecq2019high} for reconstruction of images from events. The events are compressed with PDR of $r_{4} = 1$ at 300 kbps. Fig. $5(a)$ shows the reconstruction with original events (without compression), while Fig. $5(b), (c)$ and $(d)$ shows the reconstruction for $T_{bin} = 24, 16$ and $8$. The PSNR and SSIM are computed with respect to the original reconstructed image, Fig. $5(a)$. Both PSNR and SSIM metrics increases with the increase of $T_{bin}$, indicating preference of higher $T_{bin}$ for higher image reconstruction quality, with distorted event timestamps closer to actual timestamp.

\section{Conclusion and Discussion}
The paper proposes a novel lossy event compression algorithm in the spatio-temporal domain based on Poisson disk sampling and time aggregation, which achieves state-of-the-art compression performance. The algorithm uses a QT segmentation of the intensity frames which provides priority regions. We show effectiveness of PDS-LEC in different experiments\footnote{The readers are encouraged to read the supplementary material.}: temporal, spatial, and spatio-temporal binning with comparisons and application to image reconstruction.

{\small
\bibliographystyle{ieee_fullname}
\bibliography{egbib}

\begin{thebibliography}{10}\itemsep=-1pt

\bibitem{AER_website}
\small\url{https://inivation.com/support/software/fileformat/} Last accessed on
  April 30, 2020.

\bibitem{Zstandard}
\small\url{https://facebook.github.io/zstd/} Last accessed on September 27,
  2020.

\bibitem{LZMA}
\small\url{https://www.7-zip.org/sdk.html} Last accessed on April 30, 2020.

\bibitem{snappy}
\small\url{http://google.github.io/snappy/} Last accessed on September 27,
  2020.

\bibitem{gef-dataset}
\small\url{https://sites.google.com/view/guided-event-filtering} Last accessed
  on April 30, 2020.

\bibitem{brotli}
Jyrki Alakuijala, Andrea Farruggia, Paolo Ferragina, Eugene Kliuchnikov, Robert
  Obryk, Zoltan Szabadka, and Lode Vandevenne.
\newblock Brotli: A general-purpose data compressor.
\newblock {\em ACM Transactions on Information Systems (TOIS)}, 37(1):1--30,
  2018.

\bibitem{Banerjee_EUSIPCO_2019}
Srutarshi Banerjee, Juan~G. Serra, Henry~H. Chopp, Oliver Cossairt, and
  Aggelos~K. Katsaggelos.
\newblock An adaptive video acquisition scheme for object tracking.
\newblock In {\em 2019 27th European Signal Processing Conference (EUSIPCO)},
  pages 1--5. IEEE, 2019.

\bibitem{bi2018spike}
Zhichao Bi, Siwei Dong, Yonghong Tian, and Tiejun Huang.
\newblock Spike coding for dynamic vision sensors.
\newblock In {\em 2018 Data Compression Conference}, pages 117--126. IEEE,
  2018.

\bibitem{sprintz}
Davis Blalock, Samuel Madden, and John Guttag.
\newblock Sprintz: Time series compression for the internet of things.
\newblock {\em Proceedings of the ACM on Interactive, Mobile, Wearable and
  Ubiquitous Technologies}, 2(3):1--23, 2018.

\bibitem{bridson2007fast}
Robert Bridson.
\newblock Fast poisson disk sampling in arbitrary dimensions.
\newblock {\em SIGGRAPH sketches}, 10:1278780--1278807, 2007.

\bibitem{censi2014low}
Andrea Censi and Davide Scaramuzza.
\newblock Low-latency event-based visual odometry.
\newblock In {\em 2014 IEEE International Conference on Robotics and Automation
  (ICRA)}, pages 703--710. IEEE, 2014.

\bibitem{cook1986}
Robert~L Cook.
\newblock Stochastic sampling in computer graphics.
\newblock {\em ACM Transactions on Graphics (TOG)}, 5(1):51--72, 1986.

\bibitem{BlueNoise1}
Robert~L Cook.
\newblock Stochastic sampling in computer graphics.
\newblock {\em ACM Transactions on Graphics (TOG)}, 5(1):51--72, 1986.

\bibitem{zlib}
Peter Deutsch and Jean-Loup Gailly.
\newblock Zlib compressed data format specification version 3.3.
\newblock Technical report, RFC 1950, May, 1996.

\bibitem{BlueNoise5}
Alexander Dieckmann and Reinhard Klein.
\newblock Hierarchical additive poisson disk sampling.
\newblock In {\em Proceedings of the Conference on Vision, Modeling, and
  Visualization}, pages 79--87. Eurographics Association, 2018.

\bibitem{dong2018spike}
Siwei Dong, Zhichao Bi, Yonghong Tian, and Tiejun Huang.
\newblock Spike coding for dynamic vision sensor in intelligent driving.
\newblock {\em IEEE Internet of Things Journal}, 6(1):60--71, 2018.

\bibitem{floyd1976}
Robert~W Floyd.
\newblock An adaptive algorithm for spatial gray-scale.
\newblock In {\em Proc. Soc. Inf. Disp.}, volume~17, pages 75--77, 1976.

\bibitem{event-survey}
Guillermo Gallego, Tobi Delbruck, Garrick Orchard, Chiara Bartolozzi, Brian
  Taba, Andrea Censi, Stefan Leutenegger, Andrew Davison, Joerg Conradt, Kostas
  Daniilidis, et~al.
\newblock Event-based vision: A survey.
\newblock {\em arXiv preprint arXiv:1904.08405}, 2019.

\bibitem{AVS2_overview}
Wen Gao and Siwei Ma.
\newblock An overview of avs2 standard.
\newblock In {\em Advanced Video Coding Systems}, pages 35--49. Springer, 2014.

\bibitem{khan_comparison}
Nabeel Khan, Khurram Iqbal, and Maria~G Martini.
\newblock Lossless compression of data from static and mobile dynamic vision
  sensors-performance and trade-offs.
\newblock {\em IEEE Access}, 2020.

\bibitem{TALVEN}
N. {Khan}, K. {Iqbal}, and M.~G. {Martini}.
\newblock Time aggregation based lossless video encoding for neuromorphic
  vision sensor data.
\newblock {\em IEEE Internet of Things Journal}, pages 1--1, 2020.

\bibitem{khan_BW_2019}
Nabeel Khan and Maria~G. Martini.
\newblock Bandwidth modeling of silicon retinas for next generation visual
  sensor networks.
\newblock {\em Sensors (Basel)}, 19(8):1751--1777, 2019.

\bibitem{kim2016real}
Hanme Kim, Stefan Leutenegger, and Andrew~J Davison.
\newblock Real-time 3d reconstruction and 6-dof tracking with an event camera.
\newblock In {\em Eur. Conf. Comput. Vis.}, pages 349--364. Springer, 2016.

\bibitem{lagorce2014asynchronous}
Xavier Lagorce, C{\'e}dric Meyer, Sio-Hoi Ieng, David Filliat, and Ryad
  Benosman.
\newblock Asynchronous event-based multikernel algorithm for high-speed visual
  features tracking.
\newblock {\em IEEE transactions on neural networks and learning systems},
  26(8):1710--1720, 2014.

\bibitem{2020blue}
Matteo~Paolo Lanaro, H{\'e}l{\`e}ne Perrier, David Coeurjolly, Victor
  Ostromoukhov, and Alessandro Rizzi.
\newblock Blue-noise sampling for human retinal cone spatial distribution
  modeling.
\newblock {\em Journal of Physics Communications}, 4(3):035013, 2020.

\bibitem{simdbp128}
Daniel Lemire and Leonid Boytsov.
\newblock Decoding billions of integers per second through vectorization.
\newblock {\em Software: Practice and Experience}, 45(1):1--29, 2015.

\bibitem{dvs128}
Patrick Lichtsteiner, Christoph Posch, and Tobi Delbruck.
\newblock A 128$\times$128 120 db 15 $\mu$s latency asynchronous temporal
  contrast vision sensor.
\newblock {\em IEEE journal of solid-state circuits}, 43(2):566--576, 2008.

\bibitem{choi2019learning}
IS Mostafavi, Jonghyun Choi, and Kuk-Jin Yoon.
\newblock Learning to super resolve intensity images from events.
\newblock {\em IEEE Conf. Comput. Vis. Pattern Recog.}, 2020.

\bibitem{davis}
Elias Mueggler, Henri Rebecq, Guillermo Gallego, Tobi Delbruck, and Davide
  Scaramuzza.
\newblock The event-camera dataset and simulator: Event-based data for pose
  estimation, visual odometry, and slam.
\newblock {\em The International Journal of Robotics Research}, 36(2):142--149,
  2017.

\bibitem{VP9_overview}
Debargha Mukherjee, Jim Bankoski, Adrian Grange, Jingning Han, John Koleszar,
  Paul Wilkins, Yaowu Xu, and Ronald Bultje.
\newblock The latest open-source video codec vp9-an overview and preliminary
  results.
\newblock In {\em 2013 Picture Coding Symposium (PCS)}, pages 390--393. IEEE,
  2013.

\bibitem{rebecq2018emvs}
Henri Rebecq, Guillermo Gallego, Elias Mueggler, and Davide Scaramuzza.
\newblock Emvs: Event-based multi-view stereo—3d reconstruction with an event
  camera in real-time.
\newblock {\em International Journal of Computer Vision}, 126(12):1394--1414,
  2018.

\bibitem{rebecq2018esim}
Henri Rebecq, Daniel Gehrig, and Davide Scaramuzza.
\newblock Esim: an open event camera simulator.
\newblock In {\em Conference on Robot Learning}, pages 969--982, 2018.

\bibitem{rebecq2019high}
Henri Rebecq, Ren{\'e} Ranftl, Vladlen Koltun, and Davide Scaramuzza.
\newblock High speed and high dynamic range video with an event camera.
\newblock {\em IEEE Transactions on Pattern Analysis and Machine Intelligence},
  2019.

\bibitem{scheerlinck2020fast}
Cedric Scheerlinck, Henri Rebecq, Daniel Gehrig, Nick Barnes, Robert Mahony,
  and Davide Scaramuzza.
\newblock Fast image reconstruction with an event camera.
\newblock In {\em The IEEE Winter Conference on Applications of Computer
  Vision}, pages 156--163, 2020.

\bibitem{schuster1997video}
Guido~M Schuster and Aggelos~K Katsaggelos.
\newblock A video compression scheme with optimal bit allocation among
  segmentation, motion, and residual error.
\newblock {\em IEEE Transactions on Image Processing}, 6(11):1487--1502, 1997.

\bibitem{bezier_sch}
Guido~M Schuster and Aggelos~K Katsaggelos.
\newblock An optimal quadtree-based motion estimation and motion-compensated
  interpolation scheme for video compression.
\newblock {\em IEEE Transactions on image processing}, 7(11):1505--1523, 1998.

\bibitem{shedligeri2019photorealistic}
Prasan Shedligeri and Kaushik Mitra.
\newblock Photorealistic image reconstruction from hybrid intensity and
  event-based sensor.
\newblock {\em Journal of Electronic Imaging}, 28(6):063012, 2019.

\bibitem{dvs640}
Bongki Son, Yunjae Suh, Sungho Kim, Heejae Jung, Jun-Seok Kim, Changwoo Shin,
  Keunju Park, Kyoobin Lee, Jinman Park, Jooyeon Woo, et~al.
\newblock A 640$\times$480 dynamic vision sensor with a 9$\mu$m pixel and
  {300Meps} address-event representation.
\newblock In {\em IEEE International Solid-State Circuits Conference (ISSCC)},
  pages 66--67, 2017.

\bibitem{HEVC_overview}
Gary~J Sullivan, Jens-Rainer Ohm, Woo-Jin Han, and Thomas Wiegand.
\newblock Overview of the high efficiency video coding (hevc) standard.
\newblock {\em IEEE Transactions on circuits and systems for video technology},
  22(12):1649--1668, 2012.

\bibitem{digital_half}
Robert Ulichney.
\newblock {\em Digital halftoning}.
\newblock MIT press, 1987.

\bibitem{vidal2018ultimate}
Antoni~Rosinol Vidal, Henri Rebecq, Timo Horstschaefer, and Davide Scaramuzza.
\newblock Ultimate slam? combining events, images, and imu for robust visual
  slam in hdr and high-speed scenarios.
\newblock {\em IEEE Robotics and Automation Letters}, 3(2):994--1001, 2018.

\bibitem{BlueNoise3}
Florent Wachtel, Adrien Pilleboue, David Coeurjolly, Katherine Breeden, Gurprit
  Singh, Ga{\"e}l Cathelin, Fernando De~Goes, Mathieu Desbrun, and Victor
  Ostromoukhov.
\newblock Fast tile-based adaptive sampling with user-specified fourier
  spectra.
\newblock {\em ACM Transactions on Graphics (TOG)}, 33(4):1--11, 2014.

\bibitem{eventHDR2019}
Lin Wang, Yo-Sung Ho, Kuk-Jin Yoon, et~al.
\newblock Event-based high dynamic range image and very high frame rate video
  generation using conditional generative adversarial networks.
\newblock In {\em IEEE Conf. Comput. Vis. Pattern Recog.}, pages 10081--10090,
  2019.

\bibitem{wang2020eventsr}
Lin Wang, Tae-Kyun Kim, and Kuk-Jin Yoon.
\newblock Eventsr: From asynchronous events to image reconstruction,
  restoration, and super-resolution via end-to-end adversarial learning.
\newblock {\em IEEE Conf. Comput. Vis. Pattern Recog.}, 2020.

\bibitem{gef}
Zihao~Winston Wang, Peiqi Duan, Oliver Cossairt, Aggelos Katsaggelos, Tiejun
  Huang, and Boxin Shi.
\newblock Joint filtering of intensity images and neuromorphic events for
  high-resolution noise-robust imaging.
\newblock In {\em IEEE Conf. Comput. Vis. Pattern Recog.}, 2020.

\bibitem{ed-vfs}
Zihao~Winston Wang, Weixin Jiang, Kuan He, Boxin Shi, Aggelos Katsaggelos, and
  Oliver Cossairt.
\newblock Event-driven video frame synthesis.
\newblock In {\em Proc. of the IEEE International Conference on Computer Vision
  (ICCV) Workshops}, 2019.

\bibitem{BlueNoise2}
Li-Yi Wei.
\newblock Parallel poisson disk sampling.
\newblock {\em ACM Transactions on Graphics (TOG)}, 27(3):1--9, 2008.

\bibitem{H_264}
T. {Wiegand}, G.~J. {Sullivan}, G. {Bjontegaard}, and A. {Luthra}.
\newblock Overview of the h.264/avc video coding standard.
\newblock {\em IEEE Transactions on Circuits and Systems for Video Technology},
  13(7):560--576, 2003.

\bibitem{xu2019eventcap}
Lan Xu, Weipeng Xu, Vladislav Golyanik, Marc Habermann, Lu Fang, and Christian
  Theobalt.
\newblock Eventcap: Monocular 3d capture of high-speed human motions using an
  event camera.
\newblock {\em IEEE Conf. Comput. Vis. Pattern Recog.}, 2020.

\bibitem{BlueNoise_survey}
Dong-Ming Yan, Jian-Wei Guo, Bin Wang, Xiao-Peng Zhang, and Peter Wonka.
\newblock A survey of blue-noise sampling and its applications.
\newblock {\em Journal of Computer Science and Technology}, 30(3):439--452,
  2015.

\bibitem{BlueNoise4}
Cem Yuksel.
\newblock Sample elimination for generating poisson disk sample sets.
\newblock In {\em Computer Graphics Forum}, volume~34, pages 25--32. Wiley
  Online Library, 2015.

\bibitem{zhu2017tracking}
Alex~Zihao Zhu, Nikolay Atanasov, and Kostas Daniilidis.
\newblock Event-based feature tracking with probabilistic data association.
\newblock In {\em 2017 IEEE International Conference on Robotics and Automation
  (ICRA)}, pages 4465--4470. IEEE, 2017.

\bibitem{zhu2018multivehicle}
Alex~Zihao Zhu, Dinesh Thakur, Tolga {\"O}zaslan, Bernd Pfrommer, Vijay Kumar,
  and Kostas Daniilidis.
\newblock The multivehicle stereo event camera dataset: An event camera dataset
  for 3d perception.
\newblock {\em IEEE Robotics and Automation Letters}, 3(3):2032--2039, 2018.

\bibitem{LZ77}
Jacob Ziv and Abraham Lempel.
\newblock A universal algorithm for sequential data compression.
\newblock {\em IEEE Transactions on information theory}, 23(3):337--343, 1977.

\end{thebibliography}
}

\section{Supplementary Material}
\subsection{Optimized QT based on rate distortion of Images}

Quadtree (QT) may be generated in several ways depending on the system and application. In some works in literature \cite{Banerjee_EUSIPCO_2019}, the system is developed as a Host-Chip communication problem in bandwidth-constrained environment, where the QT is developed to compress the grayscale intensity frames. In this work, we use similar system setting in order to generate the QT based on the grayscale intensity frames. For a frame $f_t$, we have a QT segmentation, skip / acquire modes for the leaves, and values for the leaves of acquire modes, denoted by $S_t$, $Q_t$, and $V_t$, respectively. These are used to reconstruct the frame $\hat{f}_t$ in a remote host, which is distorted version of $f_t$. The previously reconstructed frame $\hat{f}_{t-1}$ is used to copy the values in the skip leaves of $\hat{f}_t$, while $V_t$ provides the acquire values of the leaves in $S_t$.

The full resolution frame $f_{t+1}$ is acquired at time $t+1$ from the imager. The frames $f_{t+1}$ and $\hat{f}_t$ are inputs to the Viterbi Optimization algorithm \cite{schuster1997video}, which provides the optimal QT structure $S_{t+1}$ and skip-acquire modes $Q_{t+1}$ subject to the bandwidth constraint $B$ (expressed as maximum bit rate $R_{max}$). The Viterbi optimization provides a trade-off between the frame distortion and frame bit rate. This is done by minimizing the frame distortion $D$ over the leaves of the QT $\textbf{x}$ subject to a given maximum frame bit rate $R_{max}$. The reconstructed frame $\hat{f}_{t}$ along with frame $f_{t+1}$ is used to compute the distortion at frame $t+1$.

The optimization is formulated as follows,
\begin{align}\tag{S1}
\label{eqn:eqlabel1}
   \arg\min_{\textbf{x}} & {~D(\textbf{x})}, \\
   \text{s. t. } & {R(\textbf{x}) \leq R_{max}} \nonumber
\end{align}

The distortion for each node of the QT, \textbf{x}, is based on the skip-acquire acquisition mode $Q_t$ of that node. If a particular node $\hat{x}_t$ of a reconstructed frame at time $t$ is skip, the distortion with respect to the new node at time $t+1$, $x_{t+1}$, is given by

\begin{align}\tag{S2}
\label{eqn:eqlabel2}
    D_s = | x_{t+1} - \hat{x}_t |,
\end{align}

On the contrary, if the node is an acquire, the distortion is proportional to the standard deviation $\sigma$ as shown in Eqn. \ref{eqn:eqlabel3}:

\begin{align}\tag{S3}
\label{eqn:eqlabel3}
    D_a = \sigma \times 4^{N-n},
\end{align}
where $N$ is the maximum depth of the QT and $n$ is the level of the QT where distortion is computed. The root and the most subdivided level is defined to be in level $0$ and $N$ respectively. The total distortion is therefore defined as:

\begin{align}\tag{S4}
\label{eqn:eqlabel4}
    D = D_s + D_a,
\end{align}

\begin{figure}[t]
    \centering
    \includegraphics[width=0.8\linewidth]{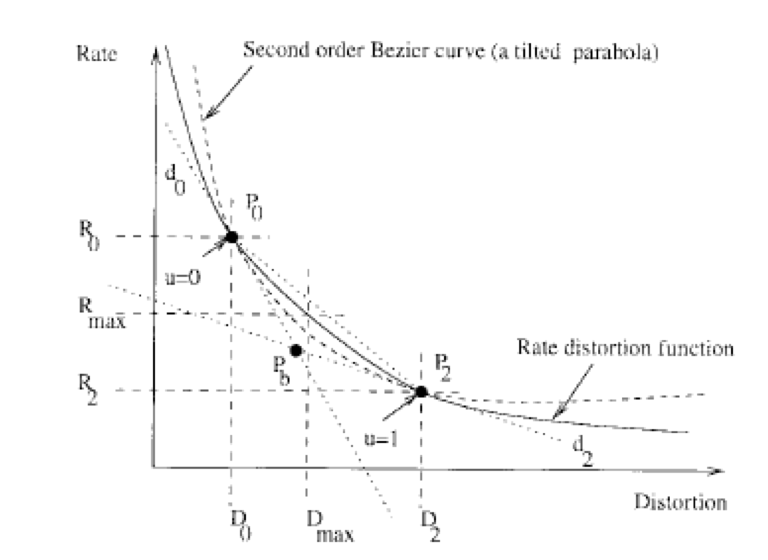}
    \caption{Rate Control using Bezier Curve \cite{bezier_sch}}
    \label{fig:bezier_rc}
\end{figure}

The constrained discrete optimization of Eqn. \ref{eqn:eqlabel1} is solved using Lagrangian relaxation, leading to solutions in the convex hull of the rate-distortion curve \cite{bezier_sch}. The Lagrangian cost function is of the form:

\begin{align}\tag{S5}
\label{eqn:eqlabel5}
    J_{\lambda} (\textbf{x}) = D(\textbf{x})+
    \lambda R(\textbf{x}),
\end{align}
where $\lambda \geq 0$ is a Lagrangian multiplier. It has been shown that if there is a $\lambda^*$ such that

\begin{align}\tag{S6}
\label{eqn:eqlabel6}
   \textbf{x}^* = \arg\min_{\textbf{x}}  {J_{\lambda^*}(\textbf{x})},
\end{align}
which leads to $R(\textbf{x}^*) = R_{max}$ (within a tolerance), then $\textbf{x}^*$ is the optimal solution to Eqn. \ref{eqn:eqlabel1}. In order to compress the grayscale intensity frames at desired bit-rate in an optimal manner, Lagrangian multiplier ($\lambda$) in Eqn. \ref{eqn:eqlabel5} is adjusted in each frame to achieve the desired bit rate. The optimal $\lambda^{*}$ is computed by a convex search in the Bezier curve \cite{bezier_sch}. A sample curve is shown in Fig. \ref{fig:bezier_rc}. In this work, for compressing the events, we consider the events in both the skip-acquire regions of the QT and compress the events depending on the QT block size. However, depending on the application, events in the acquire regions of the QT can be only compressed, while other events can be discarded.

\captionsetup{belowskip=-10pt, aboveskip=0pt}

\subsection{Comparative Performance}

In addition to the summarized results in Section $6$ of the main manuscript, we present here additional comparison results using DAVIS dataset \cite{davis} sequences: Urban, Shapes, Dynamic, Running1 and Running2. We use our PDS-LEC algorithm with quad tree (QT) generated for $0.1$ Mbps and $0.3$ Mbps for Poisson Disk Radius (PDR) of $r_4 = 1$ and $2$. In addition, the comparison of our algorithm with $50 \%$ random event sampling and reduced set of events generated by varying the thresholds to $0.25$, $0.30$ and $0.35$ generated using \cite{rebecq2018esim} is shown in Table \ref{tab:comp_lossy}. These are referred to as TH25, TH30 and TH35 in Table \ref{tab:comp_lossy}. We use the same compression strategy as mentioned in Section $6$ of the main manuscript. We compare compression ratio (CR), PSNR, SSIM and $T_{error}$ metrics for comparing the performance of the PDS-LEC with other reduced event generation methods. CR, PSNR, SSIM and $T_{error}$ is computed based on the original events provided in the DAVIS dataset \cite{davis}. Clearly TH25, TH30 and TH35 have lower PSNR, SSIM, CR and higher $T_{error}$ compared to results generated by PDS-LEC with almost all cases with different PDR and QT generated from different bit rates (PL1r100, PL2r100, PL1r300, PL2r300) as detailed in Table \ref{tab:comp_lossy}. This is seen irrespective of the different temporal binning $T_{bin}$ in the event compression algorithm. However, when considering the results of only TH25, TH30 and TH35, it is seen that the PSNR, SSIM and $T_{error}$ values are highest for TH35 compared to TH25 and TH30. This is an interesting observation, which highlights the fact that at higher contrast threshold (CT), events are fired which identify the significant edges of the scene.

However, for the cases with different CT (TH25, TH30 and TH35), the CR, PSNR, SSIM and $T_{error}$ are worse for all $T_{bins}$ compared to PDS-LEC, thereby suggesting significant difference in structure from the original event DAVIS \cite{davis} sequence. It is also observed for these cases, that for $T_{bin} = 5, 10$ and $20$ ms, the $T_{error}$ is lower than $T_{bin} = 1, 40$ ms. For $T_{bin} = 1$ ms, there are fewer events which are aggregated. This suggest that the events generated by varying CT has less structural similarity in time compared to the original events, while progressively aggregating events temporally generates an event volume which has closer resemblance to original event data. However, for $T_{bin} = 40$ ms, $T_{error}$ is higher as all event timestamps are basically quantized into $40$ ms, which increases difference between original event timestamps and quantized timestamps.

\begin{table*}[htbp]
\renewcommand\thetable{S1} 
\begin{center}
\centerline{
\resizebox{\textwidth}{!}{
    \begin{tabular}{|c|c|c|c|c|c|c|}
    \hline
    Sequence & method & 1 ms & 5 ms & 10 ms & 20 ms & 40 ms\\
    \hline 
    & & \multicolumn{5}{|c|}{(CR / PSNR / SSIM / $T_{error}$)}\\
    \hline
    \hline
    \multirow{8}{*}{Urban} & Rn & 63.35 / 53.01 / 0.9952 / 2.100 & 64.53 / 53.04 / 0.9952 / 2.111 & 66.11 / 53.32 / 0.9951 / 2.146 & 69.89 / 53.17 / 0.9949 / 2.313 & 76.06 / 53.72 / 0.9947 / 2.970\\
    & PL1r100 & 38.34 / 65.33 / 0.9987 / 1.172 & 56.41 / 59.61 / 0.9949 / 1.902 & 71.24 / 58.13 / 0.9929 / 2.192 & 91.98 / 57.21 / 0.9913 / 2.468 & 119.83 / 55.98 / 0.9899 / 2.970\\
    & PL2r100 & 45.22 / 61.72 / 0.9970 / 1.558 & 83.10 / 56.90 / 0.9911 / 2.284 & 116.22 / 55.72 / 0.9889 / 2.515 & 163.87 / 55.10 / 0.9876 / 2.689 & 231.97 / 54.53 / 0.9864 / 2.970\\
    & PL1r300 & 35.09 / 68.22 / 0.9994 / 0.929 & 46.17 / 62.08 / 0.9969 / 1.627 & 54.70 / 60.23 / 0.9954 / 1.935 & 66.15 / 58.92 / 0.9941 / 2.285 & 81.02 / 57.92 / 0.9930 / 2.970\\
    & PL2r300 & 38.70 / 64.89 / 0.9984 / 1.240 & 60.22 / 59.19 / 0.9939 / 1.997 & 78.00 / 57.68 / 0.9917 / 2.277 & 102.42 / 56.63 / 0.9901 / 2.540 & 134.76 / 55.93 / 0.9889 / 2.970\\
    & TH25 & 30.52 / 45.01 / 0.9813 / 2.906 & 36.21 / 45.01 / 0.9813 / 2.683 & 38.15 / 45.01 / 0.9813 / 2.466 & 43.90 / 45.01 / 0.9813 / 2.381 & 58.13 / 45.01 / 0.9813 / 2.970\\
    & TH30 & 38.67 / 45.64 / 0.9832 / 2.918 & 45.96 / 45.64 / 0.9832 / 2.738 & 48.11 / 45.64 / 0.9832 / 2.551 & 53.46 / 45.64 / 0.9832 / 2.445 & 68.08 / 45.64 / 0.9832 / 2.970\\
    & TH35 & 47.60 / 46.08 / 0.9840 / 2.927 & 56.61 / 46.08 / 0.9840 / 2.782 & 59.22 / 46.08 / 0.9840 / 2.625 & 64.31 / 46.08 / 0.9840 / 2.514 & 78.91 / 46.08 / 0.9840 / 2.970\\
\hline

\multirow{8}{*}{Shapes} & Rn & 61.11 / 27.79 / 0.9470 / 1.352 & 64.36 / 27.62 / 0.9458 / 1.360 & 73.38 / 27.22 / 0.9435 / 1.384 & 89.05 / 26.68 / 0.9408 / 1.494 & 107.19 / 26.16 / 0.9386 / 1.914\\

& PL1r100 & 48.45 / 30.42 / 0.974 / 1.117 & 94.88 / 25.91 / 0.9350 / 1.549 & 128.25 / 25.05 / 0.9220 / 1.637 & 172.67 / 24.48 / 0.9140 / 1.727 & 224.08 /24.09 / 0.9090 / 1.914\\

& PL2r100 & 65.99 / 27.72 / 0.9545 / 1.369 & 161.00 / 24.50 / 0.9139 / 1.710 & 234.64 / 24.00 / 0.9044 / 1.768 & 338.39 / 23.68 / 0.8981 / 1.820 & 457.91 / 23.45 / 0.8944 / 1.914\\

& PL1r300 & 40.10 / 37.20 / 0.9906 / 0.724 & 50.04 / 31.75 / 0.9756 / 1.079 & 57.33 / 30.54 / 0.9701 / 1.183 & 71.34 / 29.71 / 0.9660 / 1.386 & 89.06 / 29.16 / 0.9635 / 1.914\\

& PL2r300 & 44.56 / 34.35 / 0.9838 / 0.914 & 59.35 / 29.56 / 0.9640 / 1.253 & 68.89 / 28.66 / 0.9576 / 1.344 & 87.15 / 28.03 / 0.9523 / 1.511 & 110.42 / 27.62 / 0.9491 / 1.914\\

& TH25 & 2.94 / 1.43 / 0.8553 / 1.246 & 11.22 / 1.43 / 0.8553 / 0.911 & 18.56 / 1.43 / 0.8553 / 0.856 & 31.63 / 1.43 / 0.8553 / 1.112 & 55.46 / 1.43 / 0.8533 / 1.914\\

& TH30 & 3.54 / 3.14 / 0.8590 / 1.265 & 11.33 / 3.14 / 0.8590 / 0.963 & 19.39 / 3.14 / 0.8590 / 0.886 & 32.40 / 3.14 / 0.8590 / 1.123 & 57.57 / 3.14 / 0.8590 / 1.914\\

& TH35 & 4.14 / 4.59 / 0.8642 / 1.307 & 12.36 / 4.59 / 0.8642 / 1.007 & 20.19 / 4.59 / 0.8642 / 0.914 & 34.22 / 4.59 / 0.8642 / 1.133 & 59.91 / 4.59 / 0.8642 / 1.914\\
\hline

\multirow{8}{*}{Dynamic} & Rn & 62.45 / 28.55 / 0.8495 / 2.211 & 63.58 / 28.45 / 0.8479 / 2.222 & 67.39 / 28.16 / 0.8434 / 2.248 & 74.97 / 27.76 / 0.8373 / 2.406 & 86.04 / 27.34 / 0.8309 / 3.128\\

& PL1r100 & 38.07 / 34.63 / 0.9717 / 1.300 & 58.36 / 28.82 / 0.8890 / 2.113 & 76.54 / 27.32 / 0.8428 / 2.374 & 105.33 / 26.23 / 0.7989 / 2.643 & 143.83 / 25.54 / 0.7650 / 3.128\\

& PL2r100 & 46.55 / 31.13 / 0.9353 / 1.765 & 92.93 / 26.52 / 0.8106 / 2.536 & 132.87 / 25.60 / 0.7651 / 2.721 & 195.59 / 25.01 / 0.7295 / 2.880 & 285.71 / 24.62 / 0.7036 / 3.128\\

& PL1r300 & 34.55 / 38.08 / 0.9860 / 0.956 & 44.49 / 31.58 / 0.9375 / 1.698 & 53.53 / 29.79 / 0.9049 / 1.978 & 67.20 / 28.51 / 0.8716 / 2.335 & 84.93 / 27.67 / 0.8448 / 3.128\\

& PL2r300 & 37.94 / 34.70 / 0.9694 / 1.299 & 57.22 / 28.93 / 0.8839 / 2.098 & 73.62 / 27.58 / 0.8399 / 2.344 & 98.35 / 26.68 / 0.8016 / 2.609 & 131.73 / 26.09 / 0.7723 / 3.130\\

& TH25 & 4.12 / 3.66 / 0.545 / 2.409 & 13.63 / 3.66 / 0.545 / 2.183 & 21.49 / 3.66 / 0.545 / 2.070 & 32.78 / 3.66 / 0.545 / 2.228 & 51.24 / 3.66 / 0.545 / 3.127\\

& TH30 & 4.97 / 5.34 / 0.5580 / 2.449 & 14.61 / 5.34 / 0.5580 / 2.229 & 23.03 / 5.34 / 0.558 / 2.122 & 35.27 / 5.34 / 0.558 / 2.257 & 55.25 / 5.34 / 0.558 / 3.127\\

& TH35 & 5.83 / 6.790 / 0.5670 / 2.551 & 14.94 /  6.790 / 0.5670 / 2.266 & 24.38 / 6.790 / 0.5670 / 2.165 & 37.79 / 6.790 / 0.5670 / 2.285 & 59.12 / 6.790 / 0.5670 / 3.128\\
\hline

\multirow{8}{*}{Running1} & Rn & 62.87 / 29.20 / 0.8361 / 2.467 & 64.14 / 29.13 / 0.8346 / 2.461 & 67.61 / 28.91 / 0.8308 / 2.467 & 75.19 / 28.51 / 0.8247 / 2.6480 & 87.49 / 28.10 / 0.8176 / 3.491\\

& PL1r100 & 44.09 / 32.49 / 0.9450 / 1.808 & 74.54 / 28.06 / 0.8457 / 2.587 & 99.74 / 26.94 / 0.7977 / 2.822 & 133.89 / 26.19 / 0.7589 / 3.047 & 179.75 / 25.70 / 0.7308 / 3.491\\

& PL2r100 & 58.51 / 29.66 / 0.893 / 2.309 & 125.89 / 26.38 / 0.768 / 2.978 & 182.73 / 25.66 / 0.7239 / 3.141 & 263.88 / 25.21 / 0.6924 / 3.276 &  373.98 / 24.95 / 0.6714 / 3.491\\

& PL1r300 & 38.38 / 35.14 / 0.9678 / 1.444 & 55.31 / 29.94 / 0.8942 / 2.234 & 68.95 / 28.52 / 0.8522 / 2.499 & 87.22 / 27.58 / 0.8165 / 2.802 & 112.18 / 26.86 / 0.7895 / 3.491\\

& PL2r300 & 45.82 / 32.08 / 0.933 / 1.930 & 82.57 / 27.62 / 0.8200 / 2.697 & 110.74 / 26.64 / 0.7727 / 2.907 & 149.38 / 26.03 / 0.7373 / 3.107 & 201.25 / 25.65 / 0.7123 / 3.490\\

& TH25 & 8.70 / 7.32 / 0.5540 / 3.213 & 20.34 / 7.32 / 0.5540 / 2.841 & 26.71 / 7.32 / 0.5540 / 2.593 & 36.95 / 7.32 / 0.5540 / 2.619 & 54.26 / 7.32 / 0.5540 / 3.491\\

& TH30 & 10.62 / 8.94 / 0.5740 / 3.235 & 23.43 / 8.94 / 0.5740 / 2.926 & 30.88 / 8.94 / 0.5743 / 2.684 & 42.15 / 8.94 / 0.5743 / 2.684 & 61.00 /8.94 / 0.5740 / 3.491\\

& TH35 & 12.59 / 10.274 / 0.5880 / 3.259 & 26.71 / 10.274 / 0.5880 / 2.990 & 35.41 / 10.274 / 0.5880 / 2.764 & 47.89 / 10.274 / 0.5880 / 2.745 & 68.68 / 10.274 / 0.5880 / 3.491\\
\hline

\multirow{8}{*}{Running2} & Rn & 63.34 / 26.54 / 0.803 / 3.372 & 64.87 / 26.40 / 0.801 / 3.349 & 69.87 /  26.14 / 0.7940 / 3.336 & 80.22 / 25.73 / 0.7850 / 3.570 & 98.28 / 25.19 / 0.7740 / 4.768\\

& PL1r100 & 45.84 / 29.20 / 0.9199 / 2.607 & 80.20 / 24.92 / 0.7936 / 3.662 & 109.25 / 23.96 / 0.7380 / 3.890 & 149.63 / 23.32 / 0.6960 / 4.152 & 208.91 / 22.85 / 0.6660 / 4.768\\

& PL1r300 & 39.91 / 31.93 / 0.9505 / 2.130 & 59.29 / 26.83 / 0.8516 / 3.168 & 74.89 / 25.62 / 0.801 / 3.464 & 96.99 / 24.79 / 0.7611 / 3.822 & 131.14 / 24.15 / 0.7303 / 4.768\\

& PL2r300 &  50.46 / 28.09 / 0.8951 / 2.869 & 93.00 / 24.31 / 0.7523 / 3.810 & 126.03 / 23.55 / 0.7003 / 4.031 & 173.41 / 23.06 / 0.6635 / 4.258 & 244.27 / 22.71 / 0.6393 / 4.768\\

& TH25 & 10.27 / 7.59 / 0.5194 / 4.396 & 26.13 / 7.59 / 0.5194 / 3.913 & 36.21 / 7.59 / 0.5194 / 3.612 & 52.94 / 7.59 / 0.5194 / 3.664 & 81.19 / 7.59 / 0.5194 / 4.768\\

& TH30 & 12.48 / 9.21 / 0.5341 / 4.425 & 29.49 / 9.21 / 0.5341 / 4.018 & 40.97 / 9.21 / 0.5341 / 3.724 & 59.06 / 9.21 / 0.5341 / 3.736 & 89.38 / 9.21 / 0.5341 / 4.768\\

& TH35 & 14.72 / 10.539 / 0.5436 / 4.458 & 33.07 / 10.539 / 0.5436 / 4.098 & 45.95 / 10.539 / 0.5436 / 3.821 & 65.63 / 10.539 / 0.5436 / 3.812 & 98.65 / 10.539 / 0.55436 / 4.768\\
\hline

\end{tabular}
}
}
\end{center}
    \caption{Comparison of Compression Ratio (CR), PSNR, SSIM and $T_{error}$ for various lossy compression techniques. Rn: Random sampling of $50 \%$ events. PL1r100: PDS-LEC with $r_4 = 1$ at 0.1 Mbps. PL2r100: PDS-LEC with $r_4 = 2$ at 0.1 Mbps. PL1r300: PDS-LEC with $r_4 = 1$ at 0.3 Mbps. PL2r300: PDS-LEC with $r_4 = 2$ at 0.3 Mbps.   TH25: Fewer events with CT = 0.25. TH30: Fewer events with CT = 0.30. TH35: Fewer events with CT = 0.35.}
    \label{tab:comp_lossy}
\end{table*}

On the other hand, keeping $50 \%$ of the events randomly (Rn) using our compression technique as mentioned in section $6$ of the main paper, the CR, PSNR, SSIM and $T_{error}$ metrics are comparable to that obtained by PDS-LEC event compression technique. However, it is seen that for $T_{bin} = 1$ ms, the CR for Rn is higher, alongwith lower SSIM, PSNR and higher $T_{error}$ than PDS-LEC. This indicates that events sampled randomly at $T_{bin} = 1$ ms has less resemblance to original events in the DAVIS dataset. For higher $T_{bins} = 5, 10, 20, 40$ ms, the CR of PDS-LEC is higher than Rn, thereby implying events less than $50 \%$ are sampled. However PSNR, SSIM and $T_{error}$ of Rn is comparable to PDS-LEC.

It is observed that for PDS-LEC at $0.1$ Mbps with $r_{4} = 1, 2$, the CR is higher than that at $0.3$ Mbps, with slightly lower SSIM, PSNR and higher $T_{error}$. Clearly, PDS-LEC offers the flexibility of varying CR, and hence PSNR, SSIM and $T_{error}$, depending on the bit rate available for transmission and / or storage by varying the PDR ($r_4$) as demonstrated throughout the paper based on priority given by QT. The algorithm performs better than fewer events generated with varying CT, and comparable with events sampled randomly at 50 $\%$. However, overall, PDS-LEC has better CR and (or) PSNR / SSIM / $T_{error}$ metrics than Rn as seen in Table \ref{tab:comp_lossy}. Morever, Rn does not offer much flexibility in terms of varying the CR based on available bit rate. Thus PDS-LEC clearly has more potential and flexibility of compressing events in a lossy manner than any other method in literature or other naiive techniques.

\end{document}